\documentclass[journal]{IEEEtran}
\usepackage[numbers]{natbib}
\usepackage{graphicx}
\usepackage{multirow}
\usepackage{epstopdf}
\usepackage{times}
\usepackage{amsmath}
\usepackage{amssymb}
\usepackage{amsfonts}
\usepackage{color}
\usepackage{subfigure}
\usepackage{caption}
\usepackage{wrapfig}

\usepackage{xcolor}
\usepackage{algorithm}
\usepackage{algorithmic}
\usepackage{rotating}
\usepackage{adjustbox,lipsum}
\usepackage{dsfont}

\usepackage{booktabs}
\usepackage{multirow}
\usepackage{array}
\usepackage{graphicx}
\usepackage{tabularx}
\usepackage{colortbl} % 导入colortbl宏包
\usepackage{soul}
\definecolor{darkgreen}{HTML}{05c880}
\usepackage[colorlinks=true, citecolor=darkgreen, linkcolor=purple]{hyperref}

\definecolor{yellow_color}{HTML}{fde724}
\definecolor{purple_color}{HTML}{AF58BA}
\definecolor{blue_color}{HTML}{009ADE}
\definecolor{morandigreen}{HTML}{9DD3A8}
\definecolor{morandired}{HTML}{D9827E}
\definecolor{morandiblue}{HTML}{A3B5C6}
\definecolor{newblue}{rgb}{0.21,0.49,0.74}
\definecolor{newgray}{rgb}{0.96,1.0,0.991} 

\newcommand{\re}{\textcolor{red}}

\newcommand{\etal}{{\em et al.}}       % et al.
\newcommand{\eg}{{\em e.g.}}           % e.g.
\newcommand{\ie}{{\em i.e.}}           % i.e.

\definecolor{lightgreen}{RGB}{97, 151, 151}
\begin{document}

\title{Patch-aware Batch Normalization for Improving Cross-domain Robustness}

\author{Lei Qi,
        Dongjia Zhao,
        Yinghuan Shi,
        Xin Geng
\thanks{The work is supported by NSFC Program (Grants No. 62206052, 62125602, 62076063), China Postdoctoral Science Foundation (No. 2024M750424, GZC20240252), Jiangsu Funding Program for Excellent Postdoctoral Talent under Grant No. 2024ZB242, and the Xplorer Prize.}
\thanks{Lei Qi, Dongjia Zhao, and Xin Geng are with the School of Computer Science and Engineering, Southeast University, and Key Laboratory of New Generation Artificial Intelligence Technology and Its Interdisciplinary Applications (Southeast University), Ministry of Education, China, 211189 (e-mail: qilei@seu.edu.cn; zhaodongjia@seu.edu.cn;  xgeng@seu.edu.cn).}
\thanks{Yinghuan Shi is with the State Key Laboratory for Novel Software Technology, Nanjing University, Nanjing, China, 210023 (e-mail: syh@nju.edu.cn).}% <-this % stops a space
\thanks{Corresponding author: Xin Geng.}
%\thanks{Corresponding author: Yang Gao.}
%\thanks{Luping Zhou is School of Electrical and Information Engineering, The University of Sydney, Sydney, Australia (e-mail: luping.zhou@sydney.edu.au).}

% <-this % stops a space
%\thanks{Manuscript received O 10, 2017; revised August 26, 2017.}
}

% line to be shifted left slightly. This is one of those ``LaTeX things''. For
% instance, ``\textbf{A} \textbf{B}'' will typeset as ``A B'' not ``AB''. To get
% ``AB'' then you have to do: ``\textbf{A}\textbf{B}''

\markboth{~}%
{Shell \MakeLowercase{\textit{et al.}}: Bare Demo of IEEEtran.cls for IEEE Journals}

\maketitle

\begin{abstract}
Despite the significant success of deep learning in computer vision tasks, cross-domain tasks still present a challenge in which the model's performance will degrade when the training set and the test set follow different distributions. Most existing methods employ adversarial learning or instance normalization for achieving data augmentation to solve this task. In contrast, considering that the batch normalization (BN) layer may not be robust for unseen domains and there exist the differences between local patches of an image, we propose a novel method called patch-aware batch normalization (PBN). 
To be specific, we first split feature maps of a batch into non-overlapping patches along the spatial dimension, and then independently normalize each patch to jointly optimize the shared BN parameter at each iteration. By exploiting the differences between local patches of an image, our proposed PBN can effectively enhance the robustness of the model's parameters.  Besides, considering the statistics from each patch may be inaccurate due to their smaller size compared to the global feature maps, we incorporate the globally accumulated statistics with the statistics from each batch to obtain the final statistics for normalizing each patch. Since the proposed PBN can replace the typical BN, it can be integrated into most existing state-of-the-art methods. Extensive experiments and analysis demonstrate the effectiveness of our PBN in multiple computer vision tasks, including classification, object detection, instance retrieval, and semantic segmentation.
\end{abstract}


% \fi
\IEEEpeerreviewmaketitle

%\vspace*{-10pt}% µ÷Õû¼ä¾à
%%%%%%%%% BODY TEXT
\section{Introduction}
\IEEEPARstart{I}{n} the last decade, deep learning has made significant progress in various applications, including image classification~\cite{DBLP:conf/cvpr/HeZRS16,li2023knowledge,zha2023boosting,tang2022learning,yan2023progressive}, object detection~\cite{DBLP:conf/nips/RenHGS15}, and semantic segmentation~\cite{DBLP:conf/eccv/ChenZPSA18,li2021ctnet}. However, most existing deep learning methods assume that the training and test sets are independent and identically distributed (IID), meaning they come from the same data distribution. This assumption is often not satisfied in real-world applications, resulting in poor performance of deep models in cross-domain tasks (\ie, the training and test sets follow different distributions). For example, in urban-scene object detection, the collected images can vary widely depending on weather conditions, making it difficult to collect diverse images that cover all conditions, especially in unpredictable real-world environments.

\begin{figure}[t]
\centering
\subfigure[Images for classification.]{
\begin{minipage}[b]{0.92\linewidth}
\centering
\includegraphics[width=\textwidth]{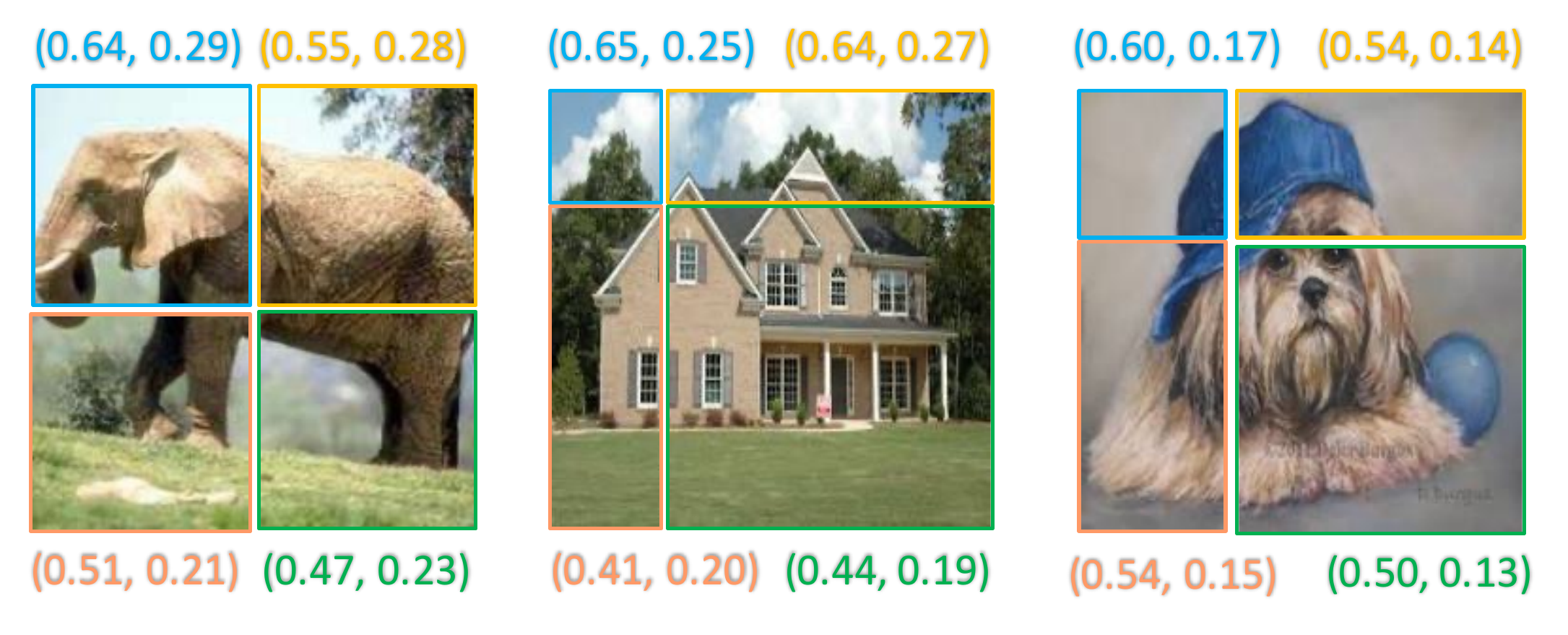}
\end{minipage}
}
\subfigure[Images for detection and segmentation.]{
\begin{minipage}[b]{0.92\linewidth}
\centering
\includegraphics[width=\textwidth]{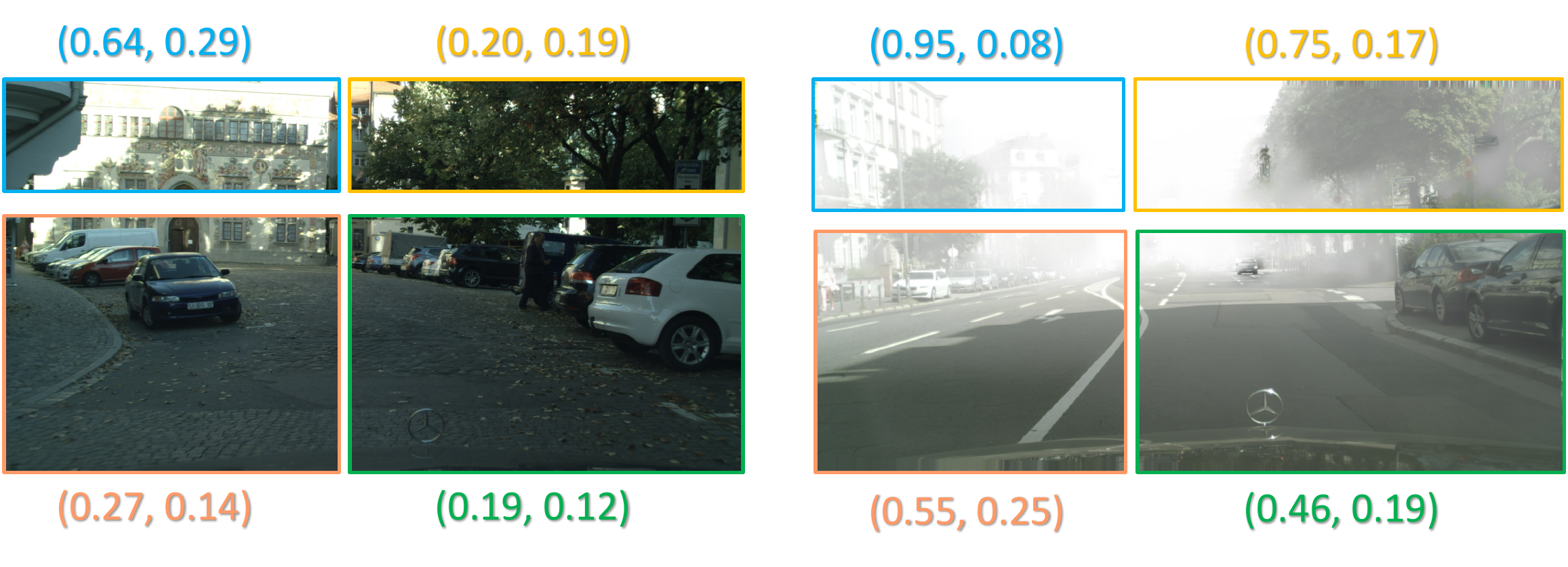}
\end{minipage}
}
\caption{The illustration of the differences between local patches within an image. We randomly divide each image into four non-overlapping patches and calculate the statistics (mean and standard deviation) of all pixels within each patch. As shown, there are discrepancies between the patches in terms of both statistics and visualization. Motivated by this observation, we propose a novel method called patch-aware batch normalization (PBN) that leverages these differences to improve cross-domain robustness.}
\label{fig06}
%\vspace*{-18pt}
\end{figure}

% \begin{figure}[t]
%   \centering
%   \begin{subfigure}{1.0\linewidth}
%   \includegraphics[width=8cm]{fig/fig6_3.pdf}%2.95
%       \caption{\scriptsize{Images for classification.}}
%     \label{fig:short-a}
%   \end{subfigure}
%   \hfill
%   \begin{subfigure}{1.0\linewidth}
%      \includegraphics[width=7.8cm]{fig/fig6_2.pdf}
%     \caption{\scriptsize{Images for detection and segmentation.}}
%     \label{fig:short-b}
%   \end{subfigure}
%   \caption{The illustration of the differences between local patches within an image. We randomly divide each image into four non-overlapping patches and calculate the statistics (mean and standard deviation) of all pixels within each patch. As shown, there are discrepancies between the patches in terms of both statistics and visualization. Motivated by this observation, we propose a novel method called patch-aware batch normalization (PBN) that leverages these differences to improve cross-domain robustness.}
%   \label{fig06}
% \end{figure}

In recent years, several domain generalization (DG) methods have been developed to address the issue of cross-domain tasks~\cite{DBLP:conf/cvpr/ZhangLLJZ22,DBLP:conf/eccv/Zhang0SG22,DBLP:conf/cvpr/ChoiJYKKC21,guo2023aloft,chen2023instance,li2023exploring,zhang2023fine}, which involves training a robust model capable of generalizing well to any unseen domains. Most of these methods are based on multiple source domains, meaning that the training set consists of samples from multiple different domains. These methods have achieved a significant improvement in the model's generalization capability due to the diverse information brought by multiple domain samples. However, it becomes even more challenging when there is only a single-source domain in the training stage, as the lack of intra-domain sample diversity makes it easier for the model to overfit to the training set. In this paper, we focus on exploring the single-source domain generalization (SDG) task.

% Recently, some domain generalization (DG) methods are developed to alleviate this issue in cross-domain tasks~\cite{DBLP:conf/eccv/HuangWXH20,DBLP:conf/cvpr/ZhangLLJZ22,DBLP:conf/eccv/Zhang0SG22,DBLP:conf/cvpr/ChoiJYKKC21}, which resorts to training a robust model to well generalize to any unseen domains. Most of these methods are based on multiple source domains, \ie, the training set consists of samples from multiple different domains. Since multiple domain samples can bring more diverse information than single domain, these methods with multiple source domains have achieved a great breakthrough on the improvement of the model's generalization capability. However, it will be more challenging when there is merely single source domain in the training stage, because it is readily to produce the overfitting to the training set when using a single domain to train the deep model.

To tackle the challenge of the single-source domain generalization (SDG), some augmentation-based methods have been developed, based on adversarial learning~\cite{DBLP:conf/cvpr/FanWKYGZ21,DBLP:conf/wacv/ChenBWS23,DBLP:conf/nips/000300M20,Cheng_2023_ICCV} or instance normalization~\cite{DBLP:conf/cvpr/NurielBW21,DBLP:conf/iclr/LiDGLSD22}, to achieve data augmentation. Additionally, an attention consistency loss is introduced to ensure alignment of class activation maps between original and augmented versions of the same training sample~\cite{DBLP:conf/cvpr/CuguMCA22}. Moreover, some multi-source domain generalization methods, such as Exact Feature Distribution Matching~\cite{DBLP:conf/cvpr/ZhangLLJZ22} and Cross-Domain Ensemble Distillation~\cite{DBLP:conf/eccv/LeeKK22}, can also be applied to the single-domain task. In contrast to these existing methods, our method focuses on addressing the issue from the perspective of batch normalization (BN).

We have noticed that there are differences between local patches in an image, as shown in Fig.~\ref{fig06}. To be specific, we randomly split an image into four non-overlapping patches and compute the statistics, such as mean and standard deviation, for all pixels in each patch. Previous studies have shown that feature statistics can carry style information for an image~\cite{DBLP:conf/iccv/HuangB17,DBLP:conf/cvpr/GatysEB16}. As shown in Fig.~\ref{fig06} (b) on the right, the top-left patch differs from the others in terms of both statistics and visualization. We aim to utilize these differences in local patches to enhance cross-domain robustness.

Moreover, several studies have shown that the traditional batch normalization (BN) method is not robust in handling domain shift~\cite{DBLP:conf/cvpr/ChangYSKH19,DBLP:conf/eccv/SeoSKKHH20}. BN typically use a pair of statistics to normalize all samples in a batch, and data diversity is lacking in a domain. As a result, we argue that the learnable affine transformation parameters of BN may not be robust in the SDG task.
To address this issue, we propose a novel patch-aware batch normalization (PBN) that leverages the differences between local patches in an image to improve the cross-domain robustness. To be specific, we randomly split the feature maps into multiple non-overlapping patches along the spatial dimension in the training stage and perform normalization for each patch independently. This method can better optimize the shared parameter of the normalization layer. Meanwhile, by leveraging the diverse information from PBN, other parameters in the model also become more robust. Besides, the statistics from each patch may not be accurate due to their smaller size compared to the global feature maps. 

To overcome this, we combine the globally accumulated statistics with the statistics from each batch to obtain the final statistics for normalizing each patch. It's worth noting that the scheme for the accumulated statistics is the same as that used in typical BN, \ie, it is accumulated by the global feature maps. Since our method is an improved BN scheme for the SDG task, it can be combined with most existing state-of-the-art (SOTA) methods. We conduct extensive experiments on multiple tasks, including classification, object detection, instance retrieval, and semantic segmentation, which verify the effectiveness of the proposed method. Furthermore, ablation studies confirm the efficacy of each component of our patch-aware batch normalization in different tasks.

Our method primarily leverages the diversity in the data distribution between patches of an image. This can be considered as some prior knowledge of each image itself. Compared to adversarial learning methods, the diversity information in our approach is not acquired through learning, and most adversarial learning methods aim to learn diversity from a vast unknown space, making it difficult to achieve an optimal solution and requiring a balance between preserving semantic information and enhancing information diversity. Additionally, regarding instance normalization methods, the literature~\cite{jin2021style} has pointed out that they lose semantic information. In contrast, BN has a good ability to retain semantic information. Our method further explores diversity while maintaining the ability to retain semantic information, resulting in better model generalization performance. %Furthermore, since our approach is an improvement based on BN, it can be applied to any backbone network containing BN layers by replacing BN with PBN. Therefore, in our experiments, we also integrate our method with other normalization methods that are not BN-based.

\section{Related work}\label{s-related}

In this section, we will provide a review of the most relevant studies to our proposed method, including works on multi-source and single-source domain generalization, as well as normalization methods for generalization.

\subsection{Domain Generalization}

Domain generalization is a challenging task in which the goal is to train a model on source domains and ensure that it can generalize well to unseen target domains. Recent studies have made significant progress in this area, including augmentation-based methods, learning domain-invariant features, and meta-learning-based methods. To address the diversity of training samples, \textit{some augmentation schemes} have been proposed to enrich training data~\cite{DBLP:conf/eccv/HuangWXH20,DBLP:conf/cvpr/ZhangLLJZ22,guo2023domaindrop,qi2024normaug,wang2022feature,qi2022novel}. From an optimization perspective, Cha \etal~\cite{DBLP:conf/nips/ChaCLCPLP21} theoretically show that finding flat minima leads to a smaller domain generalization gap. Thus, a simple yet effective method called stochastic weight averaging densely has been developed to find flat minima. Lee \etal~\cite{DBLP:conf/eccv/LeeKK22} put forward another simple yet effective method called cross-domain ensemble distillation, which \textit{learns domain-invariant features} while encouraging the model to converge to flat minima, recently found to be a sufficient condition for domain generalization. Moreover, \textit{meta-learning} has also been employed to address the DG task, which mimics the source domain and the unseen target domain using the meta-train set and the meta-test set in the training process~\cite{DBLP:conf/eccv/Zhang0SG22}. The domain-invariant feature is also beneficial for the model's generalization~\cite{DBLP:conf/eccv/MengLCYSWZSXP22,yan2023video,ding2022domain,Lin_2023_CVPR}.

\subsection{Single-source Domain Generalization}
Single-source domain generalization is a challenging task where models trained with data from only one domain are required to perform well on many unseen domains. To address this, various augmentation methods have been proposed, including both image-level~\cite{DBLP:conf/iccv/WangLQHB21,DBLP:conf/cvpr/CuguMCA22} and feature-level~\cite{DBLP:conf/cvpr/QiaoZP20,DBLP:conf/nips/000300M20,DBLP:conf/nips/VolpiNSDMS18,DBLP:conf/wacv/ChenBWS23} techniques. For instance, Chen \etal~\cite{DBLP:conf/wacv/ChenBWS23}  propose a center-aware adversarial augmentation technique that expands the source distribution by altering the source samples to push them away from class centers via an angular center loss.
In single-source DG object detection, Wu~\etal~\cite{DBLP:conf/cvpr/WuD22} present a method to disentangle single-domain generalized object detection in urban scene via cyclic-disentangled self-distillation, which learns domain-invariant features without the need for domain-related annotations.
In contrast to these methods, Wan~\etal~\cite{DBLP:conf/cvpr/WanSZY0GH022} decompose convolutional features of images into meta-features, which are defined as universal and basic visual elements for image representations.

\subsection{Normalization for Generalization}

%Exact Feature Distribution Matching for Arbitrary Style Transfer and Domain Generalization
% Zhang \etal~\cite{DBLP:conf/cvpr/ZhangLLJZ22}
% Arbitrary style transfer (AST) and domain generalization (DG) are important yet challenging visual learning tasks, which can be cast as a feature distribution matching problem. With the assumption of Gaussian feature distribution, conventional feature distribution matching methods usually match the mean and standard deviation of features. However, the feature distributions of real-world data are usually much more complicated than Gaussian, which cannot be accurately matched by using only the first-order and second-order statistics, while it is computationally prohibitive to use high-order statistics for distribution matching. In this work, we, for the first time to our best knowledge, propose to perform Exact Feature Distribution Matching (EFDM) by exactly matching the empirical Cumulative Distribution Functions (eCDFs) of image features, which could be implemented by applying the Exact Histogram Matching (EHM) in the image feature space. Particularly, a fast EHM algorithm, named Sort-Matching, is employed to perform EFDM in a plug-and-play manner with minimal cost. The effectiveness of our proposed EFDM method is verified on a variety of AST and DG tasks, demonstrating new state-of-the-art results.

%%%%%%%=============AdaIn==============

%Permuted AdaIN: Reducing the Bias Towards Global Statistics in Image Classification
Recent research has shown that the statistics of instance normalization (IN) represent the style of images. Therefore, some methods introduce noise information into IN to enhance its robustness~\cite{DBLP:conf/cvpr/NurielBW21,DBLP:conf/iclr/LiDGLSD22}. Li~\etal~\cite{DBLP:conf/iclr/LiDGLSD22} improve network generalization ability by modeling domain shift uncertainty with synthesized feature statistics during training. Tang \etal~\cite{DBLP:conf/iccv/0001GZZ0M21} introduce CrossNorm and SelfNorm. CrossNorm exchanges channel-wise mean and variance between feature maps to increase the training distribution, while SelfNorm uses attention to recalibrate statistics to bridge gaps between training and test distributions. Although CrossNorm and SelfNorm explore different directions in statistics usage, they can complement each other. Zhou \etal~\cite{DBLP:conf/iclr/ZhouY0X21} propose a novel method based on probabilistically mixing instance-level feature statistics of training samples across source domains.

Several domain generalization methods have been developed based on batch normalization. Seo~\etal~\cite{DBLP:conf/eccv/SeoSKKHH20} use multiple normalization layers and learn separate affine parameters per domain. They normalize activations for each domain by taking a weighted average of batch and instance normalization statistics. Huang \etal~\cite{DBLP:conf/cvpr/HuangZWLL22} show that estimation shift can accumulate due to the stack of batch normalization layers in a network, which can hurt test performance. To address this issue, they propose a batch-free normalization method to prevent the accumulation of estimation shift. Fan \etal~\cite{DBLP:conf/cvpr/FanWKYGZ21} introduce a generic adaptive standardization and rescaling normalization using adversarial learning. In contrast to these methods, we propose an improved batch normalization technique that exploits the differences in local patches within an image to enhance the robustness of the BN's parameters.

Compared to other existing methods, the advantages of our proposed PBN are as follows: 1) The proposed method introduces the prior information of the diversity of local image regions into BN, which has not been explored in these existing methods. 2) Unlike other methods, our approach ensures the diversity of augmented features without introducing additional noise, effectively preserving the information. 3) Our method can also be combined with other augmentation methods, further enhancing the model's generalization performance.
\section{Method}\label{s-framework}
In this paper, to improve the robustness of cross-domain models, we propose a novel method called patch-aware batch normalization (PBN) to address the overfitting to the source domain via exploiting the differences between local patches within an image. In this section, we will first review the basics of BN and then introduce PBN in detail.
\subsection{Background}
Here, we will review the conventional batch normalization (BN)~\cite{DBLP:conf/icml/IoffeS15}. First, we define feature maps of a batch as $f\in \mathbb{R}^{N\times C \times H \times W}$, where $N$ and $C$ are the number of samples and channels, and $H$ and $W$ are the height and the width of feature maps. In general, BN leverages a global statistics of a batch to normalize all samples at each iteration, which can be defined as:
\begin{equation}
  \begin{aligned}
  &{\rm BN}(f)= \gamma \frac{f-\mu}{\sigma}+\beta,%~~d \in \{1, ..., D\},
  \end{aligned}
  \label{eq01}
  \end{equation}
  where  $\gamma,\beta \in \mathbb{R}^{C}$ are learnable affine transformation parameters, and $\mu, \sigma \in \mathbb{R}^{C}$ (\ie, $\mu=[\mu_1, \cdots,\mu_C]$ and  $\sigma=[\sigma_1, \cdots,\sigma_C]$) represent the channel-wise mean and standard deviation (\ie, statistics) of BN for feature maps. For statistics of the $i$-th channel are presented as:
  \begin{equation}
  \mu_i=\frac{1}{NHW}\sum_{n=1}^{N}\sum_{h=1}^{H}\sum_{w=1}^{W}f[n,i, h,w],
    \label{eq02}
  \end{equation}
  \begin{equation}
  \sigma_i=\sqrt{\frac{1}{NHW}\sum_{n=1}^{N}\sum_{h=1}^{H}\sum_{w=1}^{W}(f[n,i,h,w]-\mu_i)^2 + \epsilon},
  \label{eq03}
  \end{equation}
  where $\epsilon$ is a constant for numerical stability. In the above equation, it can be observed that BN employs the global statistics of a batch  (\ie,  a pair of $\mu$ and $\sigma$) for all samples at each iteration. As a result, the neural network could learn a fixed pattern, which can cause the final model to overfit to the source domain, especially when the training set lacks diversity. Moreover, the values of $\gamma$ and $\beta$ in Eq.~\ref{eq01} may not be robust when the test samples do not follow the same distribution as the training samples. In this paper, we propose a novel batch normalization method to address this issue.

\subsection{Patch-aware BN}~\label{PBN}
% \begin{figure}[t]
% \centering
% \subfigure[\scriptsize{Batch Normalization}]{
%     \includegraphics[width=3.95cm]{fig/fig5_1.pdf}
%     }
% %\quad
% \subfigure[\scriptsize{Patch-aware Batch Normalization}]{
%     \includegraphics[width=3.95cm]{fig/fig5_2.pdf}
% }
% \caption{Comparison between the typical batch normalization (BN) and our patch-aware batch normalization (PBN). 
% These figures show the normalization operation for the $i$-th channel. In (b), we assume that feature maps of the $i$-th channel are randomly divided into 4 patches.}\label{fig05} 
% \vspace{-15pt}
% \end{figure}

\begin{figure}[t]
\centering
\subfigure[Batch Normalization]{
\begin{minipage}[b]{0.46\linewidth}
\centering
\includegraphics[width=\textwidth]{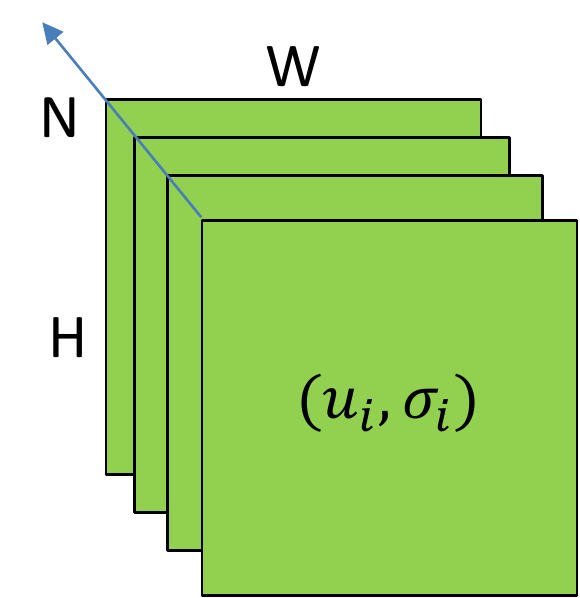}
\end{minipage}
}
\subfigure[Patch-aware Batch Normalization]{
\begin{minipage}[b]{0.46\linewidth}
\centering
\includegraphics[width=\textwidth]{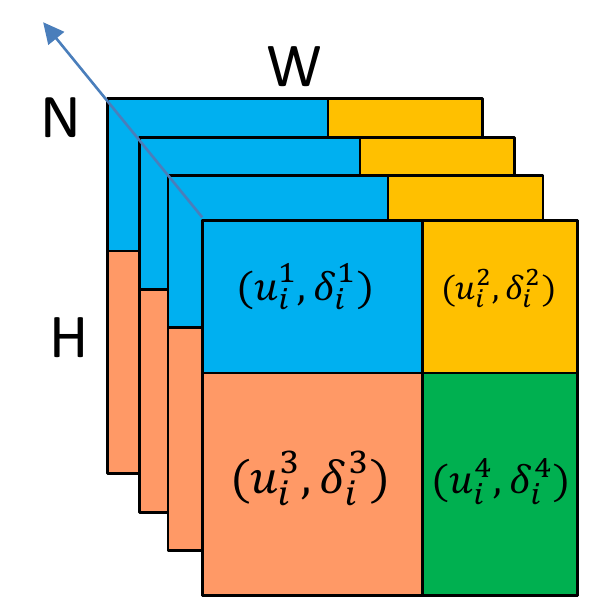}
\end{minipage}
}
\caption{Comparison between the typical batch normalization (BN) and our patch-aware batch normalization (PBN). 
These figures show the normalization operation for the $i$-th channel. In (b), we assume that feature maps of the $i$-th channel are randomly divided into 4 patches.}
\label{fig05}
%\vspace*{-15pt}
\end{figure}

% \begin{figure}[t]
%   \centering
%   \begin{subfigure}{0.49\linewidth}
%      \includegraphics[width=3.95cm]{fig/fig5_1.pdf}
%     \caption{\scriptsize{Batch Normalization}}
%     \label{fig:short-a}
%   \end{subfigure}
%   \hfill
%   \begin{subfigure}{0.49\linewidth}
%         \includegraphics[width=3.95cm]{fig/fig5_2.pdf}
%     \caption{\scriptsize{Patch-aware Batch Normalization}}
%     \label{fig:short-b}
%   \end{subfigure}
%   \caption{Comparison between the typical batch normalization (BN) and our patch-aware batch normalization (PBN). 
% These figures show the normalization operation for the $i$-th channel. In (b), we assume that feature maps of the $i$-th channel are randomly divided into 4 patches.}
%   \label{fig05}
% \end{figure}

In this paper, a novel method called Patch-aware Batch Normalization (PBN) is proposed to address the aforementioned issues of BN. PBN leverages the differences between non-overlapping patches within an image to enhance the robustness of cross-domain models, as shown in Fig.~\ref{fig06}. The discrepancy between BN and PBN is illustrated in Fig.~\ref{fig05}. In PBN, the feature maps are split into multiple non-overlapping patches, and each patch is normalized independently during the forward process. We assume that the number of patches is represented by $P$, meanwhile $H_{p}$ and $W_{p}$ denote the height and width of the $p$-th patch. We formulate PBN as follows:
\begin{equation}
  \begin{aligned}
  &{\rm PBN}(f^p)= \gamma \frac{f^p-\mu^p}{\sigma^p}+\beta,~~p \in \{1, ..., P\},
  \end{aligned}
  \label{eq04}
  \end{equation}
  where $\mu^p, \sigma^p \in \mathbb{R}^{C}$ (\ie, $\mu^p=[\mu_1^p, \cdots,\mu_C^p]$ and  $\sigma^p=[\sigma_1^p, \cdots,\sigma_C^p]$) represent the channel-wise mean and standard deviation (\ie, statistics) of the $p$-th patch within feature maps. The statistics of the $i$-th channel are written as:
  \begin{equation}
  \mu_i^p=\frac{1}{NH_{p}W_{p}}\sum_{n=1}^{N}\sum_{h=1}^{H_p}\sum_{w=1}^{W_p}f^p[n,i, h,w],
    \label{eq05}
  \end{equation}
  \begin{equation}
  \sigma_i^p=\sqrt{\frac{1}{NH_{p}W_{p}}\sum_{n=1}^{N}\sum_{h=1}^{H_p}\sum_{w=1}^{W_p}(f^p[n,i,h,w]-\mu_i^p)^2 + \epsilon}.
  \label{eq06}
  \end{equation}

Although using the method can alleviate the overfitting to the training set, $H_{p}$ and $W_{p}$ in Eqs.~\ref{eq05} and ~\ref{eq06} are smaller than $H$ and $W$ in Eqs.~\ref{eq02} and ~\ref{eq03} (refer to Fig.~\ref{fig05}), thus $\mu^p$ and $\sigma^p$ could be inaccurate when compared to the global statistics of a batch in the training course. In PBN, we further combine the globally accumulated statistics ($\hat{\mu}$ and $\hat{\sigma}$, as seen in Alg.~\ref{al01}) with the raw statistics ($\mu^p$ and $\sigma^p$) together, as inspired by~\cite{DBLP:conf/nips/KouYL020}. The final statistics of each patch can be updated as:
 \begin{equation}
 \begin{aligned}
  &\mu_i^p = \lambda \mu_i^p + (1-\lambda) \hat{\mu_i}, \\
  &\sigma_i^p = \lambda \sigma_i^p + (1-\lambda) \hat{\sigma_i}.
  \end{aligned}
  \label{eq07}
\end{equation}

% In the training stage, we randomly select the number of patches from a set $\mathbf{S}$ (\eg, $\{1,2,4\}$) for each channel respectively. To reduce the computing cost, we do not randomly split feature maps into patches independently, which requires $C$ iterations in a batch. Since the number of channels in feature maps ($C$) is often big, We group these feature maps with the same number of patches in the channel dismesion to form $G$ groups, and we randomly split feature maps into patches for each group. Therefore, we merely need $G$ iterations. The forward of the proposed PBN is described in Alg.~\ref{al01} in detail.
% During inference, we use the global accumulated statistics ($\hat{\mu}$ and $\hat{\sigma}$) to normalize feature maps.

During the training stage, we randomly select the number of patches from a set $\mathbf{S}$ (\eg, $\{1,2,4\}$)  for each channel separately. To reduce computational costs, feature maps are not randomly divided into patches independently, which requires $C$ iterations in a batch. Instead, the feature maps are grouped into $G$ groups based on the number of patches in the channel dimension, and random patch splits are applied to each group. Therefore, only $G$ iterations are required. The detailed forward process of the proposed PBN is described in Alg.~\ref{al01}. \textit{During inference, the globally accumulated statistics ($\hat{\mu}$ and $\hat{\sigma}$) are utilized to normalize feature maps, which is the same as BN.}

\begin{algorithm}[ht]
\caption{The forward process of the proposed PBN}%\newline % Ëã·¨µÄÃû×Ö
\begin{algorithmic}[1]
%\vbox{\colorbox{newgray}{\vbox{
\STATE {\bf Input:} % Ëã·¨µÄÊäÈë£¬ \hspace*{0.02in} ÓÃÀ´¿ØÖÆÎ»ÖÃ£¬Í¬Ê±ÀûÓÃ \\ ½øÐÐ»»ÐÐ
Feature maps $f$ of all samples in a batch.\\
\STATE {\bf Output:} The normalized feature maps $\hat{f}$. \\
\STATE {\bf Initialization:}  Initialize the $\lambda$ in Eq.~\ref{eq07}, the set $\mathbf{S}$ of the number of patches and the globally accumulated statistics ($\hat{\mu}$ and $\hat{\sigma}$) by the pre-trained model.\\
%\begin{algorithmic}[1]
\STATE Compute the global statistics ($\mu$ and $\sigma$) as Eqs.~\ref{eq02} and~\ref{eq03}.\\
\STATE Update the accumulated statistics ($\hat{\mu}$ and $\hat{\sigma}$) as:\\
 $\hat{\mu} = (1-m)\hat{\mu}+m\mu$ and $\hat{\sigma} = (1-m)\hat{\sigma}+m\sigma$.\\ 
 \re{\small{\textit{/* The momentum $m$ for the globally accumulated statistics is set as $0.1$ for BN in default. */}}} \\
\STATE Randomly select the number of patches ($P$) from the set $\mathbf{S}$ for each channel.\\
\STATE Group these feature maps according to the number of patches in the channel dimension to form $G$ groups. \\
\FOR {i $\in [1,...,G]$}
\STATE Randomly split feature maps into $P$ patches.
\FOR {j $\in [1,...,P]$}
\STATE Compute the raw $\mu^p$ and $\sigma^p$ as Eqs.~\ref{eq05} and~\ref{eq06}.
\STATE Update $\mu^p$ and $\sigma^p$ as Eq.~\ref{eq07}.
\STATE Conduct the normalization as Eq.~\ref{eq04}.
\ENDFOR
\ENDFOR
\STATE Regroup all patches together to obtain the normalized feature maps $\hat{f}$ according to their original positions.
%}}}
\end{algorithmic}
\label{al01}
\end{algorithm}

\textbf{Remark.} 
The proposed PBN has an advantage over the conventional BN in that it can sufficiently explore the discrepancies between local patches within an image, which allows for the information of diverse data to optimize the BN parameters. Furthermore, the generated diverse information from PBN can also enhance the robustness of other parameters in the model, especially when dealing with cross-domain shifts.
Additionally, randomly splitting all pixels of feature maps (\ie, $H\times W$ $N$-d vectors for each channel) into multiple groups and normalizing each group separately does not retain the local structural information in each group, resulting in less noticeable discrepancies. In the experimental section, we conduct experiments to confirm the importance of using the patch structure.
%could obtain robust affine transformation parameters (\ie, $\gamma$ and $\beta$ in Eq.~\ref{eq04}) when faced with cross-domain shifts. This is possible because of 

% The advantage of the proposed PBN is that our PBN can obtain the robust parameter (\ie, $\gamma$ and $\beta$ in Eq.~\ref{eq04}) when compared with the typical BN. Since there exists the discrepancy between these local patches in an image, we can exploit the discrepancy to form diverse data so as to optimize the parameter in Eq.~\ref{eq04}, which can be robust when facing the cross-domain shift. Besides, we can also randomly split the pixel (\ie, a dot in feature maps) into multiple groups and then normalize each group itself. However, the pixel based method does not keep the local structure information in each group, which could not bring the obvious discrepancy between these groups. We also validate the necessity of the patch structure in our method.

\section{Experiments}\label{s-experiment}

\subsection{Experiments on Classification}
In the classification task, we will first introduce the classification datasets and implementation details. Then, we will conduct experiments to compare our method with state-of-the-art methods. Additionally, we will perform ablation studies and further analysis to reveal the properties of our patch-aware batch normalization.
% \vspace{-10pt}
\subsubsection{Datasets and Implementation Details}
\textbf{Datasets:} In the classification task, we use PACS~\cite{Li2017DeeperBA} and CIFAR-10~\cite{krizhevsky2009learning} to perform experiments. PACS~\cite{Li2017DeeperBA} consists of four different domains: Photo, Art, Cartoon and Sketch. It contains 9,991 images with 7 object categories in total, including Photo (1,670 images), Art (2,048 images), Cartoon (2,344 images), and Sketch (3,929 images). CIFAR-10~\cite{krizhevsky2009learning} contains small $32 \times 32$ natural RGB images with respect to 10 categories,
with 50,000 training images and 10,000 testing images. In order to validate the robustness of the trained model, we
evaluate on CIFAR-10-C~\cite{DBLP:conf/iclr/HendrycksD19}, which is constructed by
corrupting the original CIFAR test sets. For the dataset, there are a total of fifteen noise, including
blur, weather, and digital corruption types, and each of them appears at five severity levels or
intensities. We follow the setting in~\cite{DBLP:conf/nips/000300M20} to report the averaged performance over
all corruptions and intensities.

\textbf{Implementation Details:} 
For the PACS dataset, we follow the same setting as in~\cite{DBLP:conf/eccv/LeeKK22,DBLP:conf/cvpr/ZhangLLJZ22}, where we use ResNet-18 as the backbone. \textit{\textbf{It is worth noting that we use the same baseline (\ie, ERM in all tables) as EFDMix}}~\cite{DBLP:conf/cvpr/ZhangLLJZ22}. We report the averaged results five times. In the case of CIFAR-10, we use Wide ResNet as the backbone as done as in~\cite{DBLP:conf/nips/000300M20}. Since our PBN can be integrated into most existing SOTA methods, we directly replace all BNs of the backbone with our PBN based on the available codes provided by their authors. Therefore, the experimental settings, including learning rate, optimizer, the number of epochs, and the final model selection, are completely consistent with the compared methods.
% \vspace{-10pt}
\subsubsection{Comparison with SOTA Methods}
\begin{figure}[t]
\begin{center}
   \includegraphics[width=0.9\linewidth]{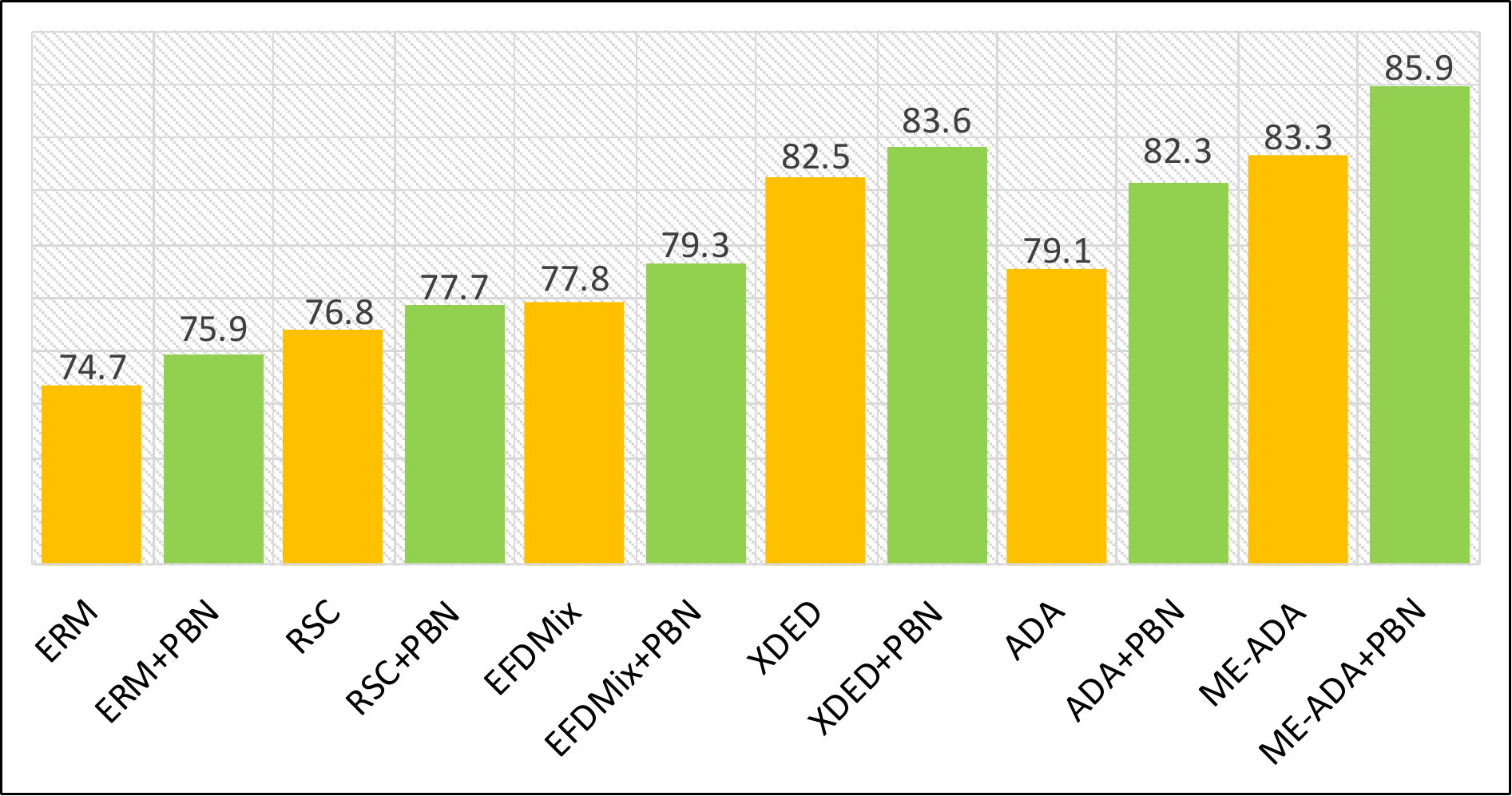}
   \caption{The experimental results of our PBN and recently SOTA methods in the ``CIFAR-10 $\to$ CIFAR-10-C'' task. Note that ``ERM'' is the baseline, \ie, the raw Wide ResNet~\cite{DBLP:conf/bmvc/ZagoruykoK16} with the cross-entropy loss.}
\label{fig04}
\end{center}
%\vspace{-25pt}
\end{figure}

We compare our method with several state-of-the-art (SOTA) methods, including RSC~\cite{DBLP:conf/eccv/HuangWXH20}, ADA~\cite{DBLP:conf/nips/VolpiNSDMS18}, ME-ADA~\cite{DBLP:conf/nips/000300M20}, EFDMix~\cite{DBLP:conf/cvpr/ZhangLLJZ22}, NP+~\cite{fan2022normalization},  ALT~\cite{Gokhale_2023_WACV} and XDED~\cite{DBLP:conf/eccv/LeeKK22}, and integrate our PBN into these methods by replacing BN of the backbone. The experimental results, reported in Tab.~\ref{tab01}, show that incorporating our method consistently improves performance across all methods. For example, compared with recent methods such as EFDMix~\cite{DBLP:conf/cvpr/ZhangLLJZ22} and XDED~\cite{DBLP:conf/eccv/LeeKK22}, using PBN can increase their performance by $+2.9\%$ (57.3 vs. 54.4) and $+2.3\%$ (68.8 vs. 66.5), respectively, demonstrating the effectiveness of our proposed PBN. We also report experimental results for the ``CIFAR-10 $\to$ CIFAR-10-C'' task, which show that integrating our method into these existing SOTA methods can also achieve further improvements, as illustrated in Fig.~\ref{fig04}.

% We compare our method with some state-of-the-art (SOTA) methods, including RSC~\cite{DBLP:conf/eccv/HuangWXH20}, ADA~\cite{DBLP:conf/nips/VolpiNSDMS18}, ME-ADA~\cite{DBLP:conf/nips/000300M20}, EFDMix~\cite{DBLP:conf/cvpr/ZhangLLJZ22} and EFDMix~\cite{DBLP:conf/cvpr/ZhangLLJZ22}. Since our PBN is plug-and-play, it can be integrated into to other methods by replacing BN. Experimental results are reported in Table~\ref{tab01}. In this PACS experiment, we use a domain as test domain, and the others are used as the training domain respectively. As seen, fusing our method into these existing methods can consistently obtain the improvement. For instance, compared with these recent methods, such as EFDMix~\cite{DBLP:conf/cvpr/ZhangLLJZ22} and EFDMix~\cite{DBLP:conf/cvpr/ZhangLLJZ22}, using PBN can increase their performance by $+2.9\%$ (57.3 vs. 54.4) and $+2.2\%$ (68.7 vs. 66.5), showing the effectiveness of the proposed PBN. In addition, we also report experimental results in the ``CIFAR-10 $\to$ CIFAR-10-C'' task, which also show that plugging our method into these existing SOTA methods can also achieve further improvement, as illustrated in Fig.~\ref{fig04}.

\begin{table}[htbp]
  \centering
  \caption{Comparison with SOTA methods on PACS. We list the result when using Photo (P),  Art (A), Cartoon (C), and Sketch (S) as the test domain respectively. For example, P denotes the averaged result of A $\to$ P, C $\to$ P and S $\to$ P. Note that ``ERM'' is the baseline, \ie, the raw ResNet-18~\cite{DBLP:conf/cvpr/HeZRS16} with the cross-entropy loss.}
  \setlength{\tabcolsep}{3.5mm}{
    \begin{tabular}{l|cccc|c}
    \toprule
    \multicolumn{1}{c|}{Method} & P & A   & C & S & Avg \\
    \midrule
    ERM   & 34.0  & 58.6  & 66.4  & 27.5  & 46.6 \\
    \rowcolor{gray!20} % 使用gray!20指定灰色的底色，20表示灰度值
    \multicolumn{1}{l|}{ERM+PBN} & \textbf{36.2}  & \textbf{60.5}  & \textbf{69.1}  & \textbf{31.5}  & \textbf{49.3} \\
    \midrule
    RSC~\cite{DBLP:conf/eccv/HuangWXH20}   & \textbf{57.3}  & 75.3  & 77.4  & 45.9  & 64.0 \\
    \rowcolor{gray!20} % 使用gray!20指定灰色的底色，20表示灰度值
    RSC+PBN & 55.2  & \textbf{76.8}  & \textbf{78.5}  & \textbf{57.0}  & \textbf{66.9} \\
    \midrule
    ADA~\cite{DBLP:conf/nips/VolpiNSDMS18} &    33.6   &   59.7    &  67.1     &  \textbf{27.3}     & 46.9 \\
    \rowcolor{gray!20} % 使用gray!20指定灰色的底色，20表示灰度值
    ADA+PBN &   \textbf{36.9}    &   \textbf{61.5}    &  \textbf{69.1}     &   26.6    & \textbf{48.5} \\
    \midrule
    ME-ADA~\cite{DBLP:conf/nips/000300M20} &   33.6    &   59.7    &   67.1    &   26.8    & 46.8 \\
    \rowcolor{gray!20} % 使用gray!20指定灰色的底色，20表示灰度值
    ME-ADA+PBN &   \textbf{37.1}    &   \textbf{61.0}    &  \textbf{68.9}     &  \textbf{28.2}     &  \textbf{48.8}\\
    \midrule
    EFDMix~\cite{DBLP:conf/cvpr/ZhangLLJZ22} & 42.5  & 63.2  & \textbf{73.9}  & 38.1  & 54.4 \\
    \rowcolor{gray!20} % 使用gray!20指定灰色的底色，20表示灰度值
   EFDMix+PBN & \textbf{49.1}  & \textbf{66.9}  & 73.0  & \textbf{40.1}  & \textbf{57.3} \\
      \midrule
    NP+~\cite{fan2022normalization} & 37.9  & 65.2  & 61.7  & 41.5  & 51.6 \\
    \rowcolor{gray!20} % 使用gray!20指定灰色的底色，20表示灰度值
    NP++PBN & \textbf{43.3}  & \textbf{67.9}  & \textbf{62.7}  & \textbf{42.0}  & \textbf{53.9} \\
      \midrule
    ALT~\cite{Gokhale_2023_WACV} & 54.6 & 74.9  & 75.5  & 51.0  & 64.0 \\
    \rowcolor{gray!20} % 使用gray!20指定灰色的底色，20表示灰度值
    ALT+PBN & \textbf{57.6}  & \textbf{76.0}  & \textbf{76.3}  & \textbf{53.5}  & \textbf{65.9} \\
    \midrule
    XDED~\cite{DBLP:conf/eccv/LeeKK22} & 59.1  & 76.5  & 77.2  & 53.1  & 66.5 \\
    \rowcolor{gray!20} % 使用gray!20指定灰色的底色，20表示灰度值
    XDED+PBN & \textbf{62.2}  & \textbf{78.3}  & \textbf{79.9}  & \textbf{54.6}  & \textbf{68.8} \\
    \bottomrule
    \end{tabular}}
  \label{tab01}%
  %\vspace{-10pt}
\end{table}%

We also compare our method with the normalization based methods, including CNSN~\cite{DBLP:conf/iccv/0001GZZ0M21}, DSU~\cite{DBLP:conf/iclr/LiDGLSD22} and 
pAdaIN~\cite{DBLP:conf/cvpr/NurielBW21}. The experimental results are listed in Tab.~\ref{tab02}. As observed in this table, our method outperforms these compared methods. For example, compared with DSU, the performance of our PBN increases it by $+2.4\%$ (49.3 vs. 46.9). It is worth noting that, our PBN sightly outperforms pAdaIN, but our method can be integrated into it and further improve its performance. Most normalization based methods aim to introduce extra and diverse information into instance normalization, while our proposed method is an improved batch normalization that can be easily integrated into existing methods by replacing batch normalization with our patch-aware batch normalization. This allows for the combination of the advantages of different normalization methods, leading to improved performance on various tasks.
\begin{table}[htbp]
  \centering
  \caption{Comparison with the normalization based methods in the image classification task on PACS. In this table, the \textbf{bold} is the best performance.}
  \setlength{\tabcolsep}{3.3mm}{
    \begin{tabular}{l|cccc|c}
    \toprule
    \multicolumn{1}{c|}{Method} & P     & A     & C     & S     & ~~Avg~~ \\
    \midrule
    CNSN~\cite{DBLP:conf/iccv/0001GZZ0M21}  &   31.6   &   \textbf{61.7}    &   66.1    &   20.2    & 44.9 \\
  DSU~\cite{DBLP:conf/iclr/LiDGLSD22} & 33.8  & 58.8  & 67.2  & 27.9  & 46.9 \\
   pAdaIN~\cite{DBLP:conf/cvpr/NurielBW21} & 35.6  & 60.5  & 68.0  & 30.6  & 48.7 \\
    \rowcolor{gray!20} % 使用gray!20指定灰色的底色，20表示灰度值
    PBN (Ours) & \textbf{36.2}  & 60.5  & \textbf{69.1}  & \textbf{31.5}  & \textbf{49.3} \\
    \midrule
    CNSN+PBN  &  \textbf{38.3}     &   \textbf{62.2}    &   65.4    &   27.9    & 48.5 \\ 
    DSU+PBN & 36.3  & 61.2  & 69.0  & 29.5  & 49.0 \\
    \rowcolor{gray!20} % 使用gray!20指定灰色的底色，20表示灰度值
    pAdaIN+PBN & 38.1  & 61.3  & \textbf{69.3}  & \textbf{29.6}  & \textbf{49.6} \\
    \bottomrule
    \end{tabular}}
  \label{tab02}%
  %\vspace{-10pt}
\end{table}%

\subsubsection{Ablation Studies and Further Analysis}
\textbf{Ablation studies.} In section~\ref{PBN}, we leverage the globally accumulated statistic to mitigate the instability of the statistics from each patch in the training stage. Here, we perform the experiment to validate the effectiveness of both the module in PBN and PBN without the module, as shown in Tab.~\ref{tab03}. As seen in this table, our method without the globally accumulated statistics (\ie, ``ERM+PBN w/o GS'' in Tab.~\ref{tab03}) can still improve the performance of the baseline (\ie, ``ERM'') by $+2.2\%$ (48.8 vs.~46.6). Furthermore, when integrating the globally accumulated statistic into PBN, our method can obtain the further improvement, which demonstrates the necessity of the the globally accumulated statistic in PBN. In addition, in order to sufficiently validate the effectiveness of each component in our method, we will also do ablation studies in the other tasks.
\begin{table}[htbp]
  \centering
  \caption{Ablation studies in the image classification task on PACS. ``w/o'' denotes ``without'',  and ``GS'' indicates the globally accumulated statistic, \ie, $\hat{\mu}$ and $\hat{\sigma}$ in Eq.~\ref{eq07}.}
    \setlength{\tabcolsep}{3.3mm}{
    \begin{tabular}{l|cccc|c}
    \toprule
    \multicolumn{1}{c|}{Method} & P     & A     & C     & S     & Avg \\
    \midrule
    ERM & 34.0  & 58.6  & 66.4  & 27.5  & 46.6 \\
    ERM+PBN w/o GS & \textbf{36.3}  & \textbf{60.5}  & \textbf{69.3}  & 29.0  & 48.8 \\
    \rowcolor{gray!20} % 使用gray!20指定灰色的底色，20表示灰度值
    ERM+PBN & 36.2  & \textbf{60.5}  & 69.1  & \textbf{31.5}  & \textbf{49.3} \\
    \bottomrule
    \end{tabular}}
  \label{tab03}
 % \vspace{-5pt}
\end{table}%

\textbf{Evaluation of different splitting schemes and the number of patches.} In this experiment, we present the results of different splitting schemes and the number of patches used on PACS, as shown in Tab.~\ref{tab04}. It's important to note that to observe the influence of different splitting schemes and patch numbers, we use the same scheme across all feature map channels. For the random splitting scheme, we ensure that each patch is not too small by randomly selecting a position between $1/3$ height (or width) and $2/3$ height (or width) for 2 and 4 patches. For 9 patches, we select two positions (\ie, $1/5$ to $2/5$ and $3/5$ to $4/5$) in both height and width dimensions. As can be seen in the table, the random splitting scheme generally outperforms the equal splitting scheme. Additionally, we find that randomly splitting feature maps into 4 patches (\ie, ``P4-random'') achieves the best results.

\begin{table}[htbp]
  \centering
  \caption{Results of different splitting schemes and the number of patches on PACS. In this table, the notation ``P2'' represents splitting feature maps into two patches. ``LR" and ``UD'' represent splitting feature maps into left-right and up-down patches, respectively. ``Equal'' indicates that all patches are of the same size, while ``random'' indicates that feature maps are randomly divided into different patches.}
  \setlength{\tabcolsep}{3.3mm}{
    \begin{tabular}{l|cccc|c}
    \toprule
    \multicolumn{1}{c|}{Method} & P     & A     & C     & S     & ~~Avg~~ \\
     \midrule
     Baseline & 34.0  & 58.6  & 66.4  & 27.5  & 46.6  \\
    \midrule
    P2-LR-equal & 34.9  & 59.5  & 68.3  & 27.6  & 47.6 \\
    P2-UD-equal & 35.5  & 60.5  & 67.7  & 27.2  & 47.7 \\
    P2-LR-random & 35.7  & 60.7  & \textbf{69.5}  & 29.0  & 48.7 \\
    P2-UD-random & 36.4  & 60.9  & 68.2  & 29.0  & 48.6 \\
    P4-equal & 35.6  & 60.4  & 68.0  & 28.8  & 48.2 \\
    \rowcolor{gray!20} % 使用gray!20指定灰色的底色，20表示灰度值
    P4-random & \textbf{36.7}  & \textbf{61.4}  & 68.9  & 29.2  & \textbf{49.1} \\
    P9-equal & 34.2  & 60.4  & 65.8  & \textbf{33.0}  & 48.4 \\
    P9-random & 34.6  & 61.1  & 66.6  & 31.3  & 48.4 \\
    \bottomrule
    \end{tabular}}
  \label{tab04}%
  %\vspace{-10pt}
\end{table}%

\setlength{\columnsep}{10pt}%
%\begin{wrapfigure}[14]{r}[1pt]{4.5cm}  %% I used 8 here, change suitably
\begin{wrapfigure}{r}{4.5cm} 
    \centering
    \includegraphics[width=4.5cm]{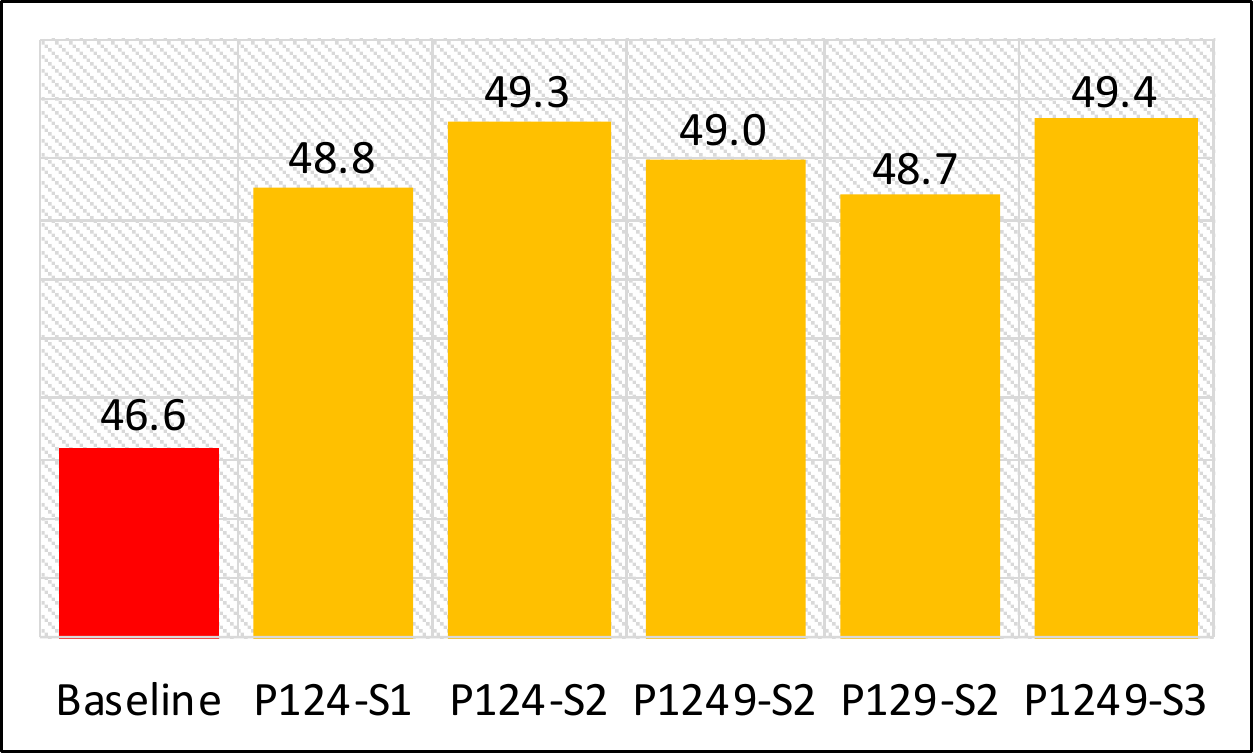}\\
     \captionof{figure}{The analysis of the randomness along the channel dimension. In this figure, ``P124-S2'' denotes that we randomly select 2 digits from $\{1,2,4\}$ as the number of patches.}
     \label{fig03}
\end{wrapfigure}
\textbf{Comparison with the pixel based scheme.} In this experiment, we further explore the necessity of using patches in our method. We conduct this experiment using a pixel based scheme, which randomly splits all pixels of a batch (\ie, $H \times W$ $N$-dimensional vectors for each channel) into multiple pixel groups. To have a fair comparison, the number of pixel groups is the same as the number of divided patches, and the size of each group is generated by the same random scheme as the patch splitting scheme (\ie, ``P4-random'' in Tab.~\ref{tab04}). The experimental results are listed in Tab.~\ref{tab05}. As shown, the patch-based scheme is necessary and meaningful in our method, while the pixel based scheme neglects the structural information in the spatial dimension. The differences across local patches can be leveraged to alleviate overfitting to the training set in cross-domain tasks.
% \vspace{-10pt}
\begin{table}[htbp]
  \centering
  \caption{Experimental results of the patch based scheme and the pixel based scheme via ``P4-random'' in Tab.~\ref{tab04} .}
    \setlength{\tabcolsep}{3.3mm}{
    \begin{tabular}{l|cccc|c}
    \toprule
    \multicolumn{1}{c|}{Method} & P     & A     & C     & S     & ~~Avg~~ \\
    \midrule
    BN (Baseline) & 34.0  & 58.6  & 66.4  & 27.5  & 46.6 \\
    Pixel & 34.6 & 60.4 & 67.9 & 24.7 & 46.9 \\
    \rowcolor{gray!20} % 使用gray!20指定灰色的底色，20表示灰度值
    Patch & \textbf{36.7}  & \textbf{61.4}  & \textbf{68.9}  & \textbf{29.2}  & \textbf{49.1} \\
    \bottomrule
    \end{tabular}}
  \label{tab05}%
  \vspace{-5pt}
\end{table}%

%\lipsum[6]
%\endgroup

\textbf{The analysis of the randomness along the channel.} 
To analyze the necessity of randomness on the channel dimension, we conduct experiments using different numbers of patch sets and randomly select numbers, as shown in Fig.~\ref{fig03}. As observed in the figure, the ``P124-S2'' setting achieve good performance, and adding the set size and randomly selected number slightly increased performance. Considering that choosing a more complex setup only brings slight improvements, we opt for a relatively simple setup, which also yields good results. Therefore, we select ``P124-S2'' as our final setting for classification.

\begin{figure}[t]
\centering
\subfigure[ ]{
\begin{minipage}[b]{0.46\linewidth}
\centering
\includegraphics[width=\textwidth]{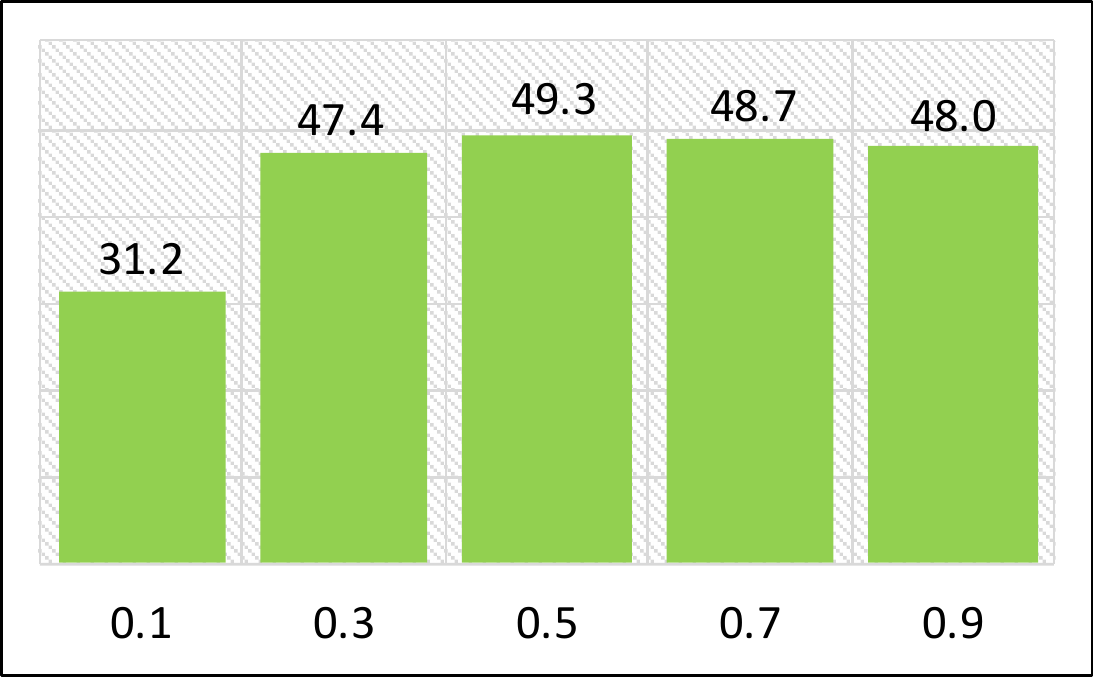}
\end{minipage}
}
\subfigure[ ]{
\begin{minipage}[b]{0.46\linewidth}
\centering
\includegraphics[width=\textwidth]{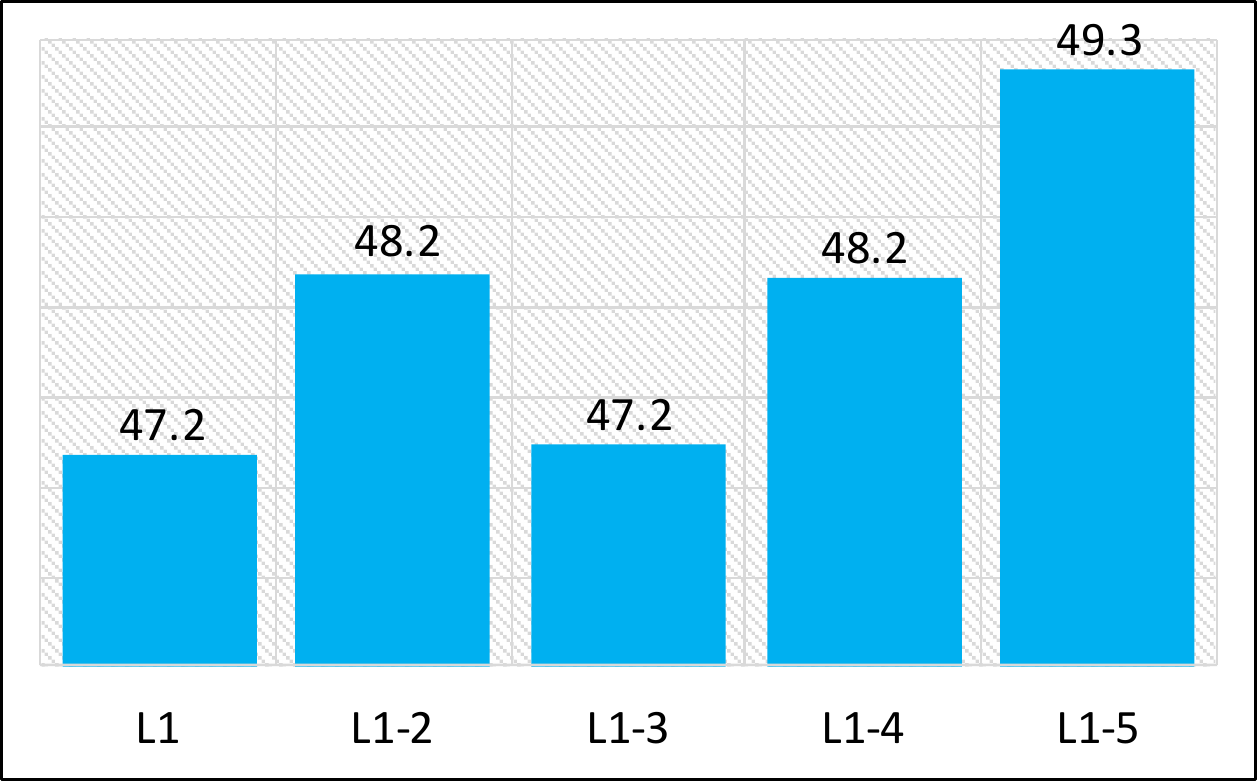}
\end{minipage}
}
\caption{The analysis of the hyper-parameter $\lambda$ in Eq.~\ref{eq07} in (a). (b) is the result of using our method at different positions of the backbone. Both (a) and (b) are conducted on PACS.}
\label{fig01}
%\vspace*{-20pt}
\end{figure}

\textbf{Evaluation of the impact of our method at different positions.} ResNet consists of five modules, \ie, a convolutional module and four residual block modules.  In Fig.~\ref{fig01} (b), we give the experimental result when using our PBN at different positions. In this figure, ``L1-5'' indicates that we replace all BNs of five modules.  As seen, when replacing all raw BNs with PBN, our method can obtain the best result on PACS. In the classification task, we utilize this setting to conduct all experiments.
% \begin{figure}[htbp]
% \centering
% \subfigure[]{
%     \includegraphics[width=3.95cm]{fig/fig1.pdf}
%     }
% %\quad
% \subfigure[]{
%     \includegraphics[width=3.95cm]{fig/fig2.pdf}
% }
% \caption{The analysis of the hyper-parameter $\lambda$ in Eq.~\ref{eq07} in (a). (b) is the experimental result of using our method at different positions of the backbone. Both (a) and (b) are conducted on PACS.}\label{fig01} 
% \vspace{-10pt}
% \end{figure}

% \begin{figure}
%   \centering
%   \begin{subfigure}{0.49\linewidth}
%        \includegraphics[width=3.95cm]{fig/fig1.pdf}
%     \caption{ }
%     \label{fig:short-a}
%   \end{subfigure}
%   \hfill
%   \begin{subfigure}{0.49\linewidth}
%      \includegraphics[width=3.95cm]{fig/fig2.pdf}
%     \caption{ }
%     \label{fig:short-b}
%   \end{subfigure}
% \caption{The analysis of the hyper-parameter $\lambda$ in Eq.~\ref{eq07} in (a). (b) is the experimental result of using our method at different positions of the backbone. Both (a) and (b) are conducted on PACS.}\label{fig01} 
% \end{figure}

\textbf{The analysis of the hyper-parameter.} 
In Eq.~\ref{eq07}, the hyper-parameter $\lambda$ is used to balance the influence of the globally accumulated statistics and the statistics of each patch. We analyze the effect of different values of $\lambda$, as shown in Fig.~\ref{fig01} (a). From the figure, it can be observed that the best performance is achieved when $\lambda$ is set to 0.5. Moreover, there is a significant decrease in performance when using only the globally accumulated statistics, as this method reduces the diversity of information.
%\begingroup

% \begin{figure}[t]
% \begin{center}
%    \includegraphics[width=0.9\linewidth]{fig/fig3.pdf}
%    \caption{The analysis of the randomness along the channel. In this figure, ``P124-S2'' denotes that we randomly select 2 digits from $\{1,2,4\}$ as the number of patches.}
% \label{fig03}
% \end{center}
% \vspace{-25pt}
% \end{figure}

% \begin{table}[t]
% \centering
% \caption{The analysis of the randomness along the channel. In this figure, ``P124-S2'' denotes that we randomly select 2 digits from $\{1,2,4\}$ as the number of patches.}
% \label{tab:select-way}
% \setlength{\tabcolsep}{4.1mm}{
% \begin{tabular}{l|cccc|c}
% \toprule
% \multicolumn{1}{c|}{Method} & P     & A     & C     & S     & Avg \\
% \midrule
% Baseline & 58.6 & 66.4 & 34.0 & 27.5 & 46.6 \\
% P124-S1 & 60.5 & 69.3 & 36.3 & 29.0 & 48.8 \\
% P124-S2 & 60.5 & 69.1 & 36.2 & 31.5 & 49.3 \\
% P1249-S2 & 61.4 & 68.4 & 35.9 & 30.3 & 49.0 \\
% P129-S2 & 61.3 & 68.0 & 35.5 & 30.1 & 48.7 \\
% P1249-S3 & 60.8 & 68.8 & 36.4 & 31.4 & 49.4 \\
% \bottomrule
% \end{tabular}}
% \vspace*{-5pt}
% \end{table}

% \begin{wrapfigure}[8]{r}{20cm}
% \begin{minipage}{0.25\textwidth}
% %\begin{figure}[t]
% \begin{center}
%    \includegraphics[width=1\linewidth]{fig/fig3.pdf}
%    \captionof{figure}{The analysis of on randomness on the channel. In this figure, ``P124-S2'' denotes that we randomly select 2 digits from $\{1,2,4\}$ as the number of patches.}
% \end{center}

% \label{fig03}
% %\end{figure}
% \end{minipage}%\hfill
% \end{wrapfigure}

%\setlength{\intextsep}{1cm}

\textbf{Results based on ResNet-50.} To further validate the effectiveness of our method, we also use ResNet-50 as the backbone to conduct the experiment on PACS. The experimental results are listed in Tab.~\ref{tab11}. For example, when integrating our PBN into EFDMix~\cite{DBLP:conf/cvpr/ZhangLLJZ22}, it can still increase $+2.8\%$ (64.0 vs. 61.2). \textit{We also conduct the ablation based on VGG~\cite{simonyan2015very}, and put the result in the supplementary materials.}
\begin{table}[htbp]
  \centering
  \caption{Experimental results of using ResNet-50 as the backbone on PACS.}
  \setlength{\tabcolsep}{3.5mm}{
    \begin{tabular}{l|cccc|c}
    \toprule
     \multicolumn{1}{c|}{Method} & P     & A     & C     & S     & Avg \\
    \midrule
    ERM & 38.0  & 63.5  & 69.2  & 31.4  & 50.5 \\
    \rowcolor{gray!20} % 使用gray!20指定灰色的底色，20表示灰度值
   ERM+PBN & \textbf{41.4}  & \textbf{67.0}  & \textbf{71.5}  & \textbf{38.0}  & \textbf{54.5} \\
    \midrule
    EFDMix~\cite{DBLP:conf/cvpr/ZhangLLJZ22} & 48.0  & 75.3  & 77.4  & 44.2  & 61.2 \\
    \rowcolor{gray!20} % 使用gray!20指定灰色的底色，20表示灰度值
    EFDMix+PBN & \textbf{55.3}  & \textbf{78.2}  & \textbf{75.9}  & \textbf{46.4}  & \textbf{64.0} \\
    \bottomrule
    \end{tabular}}
  \label{tab11}%
  \vspace{-15pt}
\end{table}%

\subsection{Generalization on Object Detection}
%\subsubsection{Datasets and Implementation Details}

\textbf{Dataset:} We employ the Diverse-Weather dataset \cite{DBLP:conf/cvpr/WuD22} to conduct the experiment in the object detection task. It consists of five scenes with different weather conditions, \ie, daytime-sunny, night-sunny, dusk-rainy, night-rainy, and daytime-foggy. Particularly, the daytime-sunny scene is the source domain for training, and the night-sunny, dusk-rainy, night-rainy, and daytime-foggy scenes are target domains for testing. For the daytime-sunny scene, 19,395 images are for training. The night-sunny, dusk-rainy, night-rainy, and daytime-foggy scenes separately include 26,158,  3,501,  2,494, and 3,775  images for test. 

\textbf{Implementation details:} 
For the Diverse-Weather dataset, we train our model on the daytime-sunny domain and evaluate it on other domains. In contrast to the classification task, the object detection task involves identifying small objects, so we retain the batch normalization (BN) layer in the last two ResNet blocks and replace the other BNs with the proposed PBNs. We implement our method based on the available codes provided by the authors of the SOTA methods and adapt them to our method.
\newcommand{\PreserveBackslash}[1]{\let\temp=\\#1\let\\=\temp}
\newcolumntype{C}[1]{>{\PreserveBackslash\centering}p{#1}}
\newcolumntype{R}[1]{>{\PreserveBackslash\raggedleft}p{#1}}
\newcolumntype{L}[1]{>{\PreserveBackslash\raggedright}p{#1}}
\begin{table}[htbp]
  \centering
  \caption{Comparison with SOTA methods in the object detection task. The model is trained on daytime-sunny and test on Night-Sunny (NS), Dusk-Rainy (DR), Night-Rainy (NR) and Daytime-Foggy (DF), respectively.}
  \setlength{\tabcolsep}{3.3mm}{
    \begin{tabular}{l|C{5mm}C{5mm}C{5mm}C{5mm}|c}
    \toprule
    \multicolumn{1}{c|}{Method} & NS & DR & NR & DF & Avg \\
    \midrule
    CycConf~\cite{DBLP:conf/iccv/0066HLY0GD21} & \textbf{45.2}  & 36.0  & 15.7  & 38.3  & 33.8 \\
    \rowcolor{gray!20} % 使用gray!20指定灰色的底色，20表示灰度值
    CycConf+PBN & 44.8  & \textbf{40.8}  & \textbf{19.6}  & \textbf{38.9}  & \textbf{36.0} \\
    \midrule
    CDSD~\cite{DBLP:conf/cvpr/WuD22}  & 37.4  & 27.2  & 12.9  & 33.3  & 27.7 \\
    \rowcolor{gray!20} % 使用gray!20指定灰色的底色，20表示灰度值
    CDSD+PBN &    \textbf{38.1}   &   \textbf{28.4}    &  \textbf{14.0}     &   \textbf{32.6}    & \textbf{28.3} \\
    \midrule
    Faster R-CNN~\cite{DBLP:conf/nips/RenHGS15} & 46.8  & 37.5  & 16.7  & 38.1  & 34.8 \\
    \rowcolor{gray!20} % 使用gray!20指定灰色的底色，20表示灰度值
    Faster R-CNN+PBN & \textbf{47.0}  & \textbf{44.3}  & \textbf{23.8}  & \textbf{40.4}  & \textbf{38.9} \\
    \bottomrule
    \end{tabular}}
  \label{tab06}%
  %\vspace{-10pt}
\end{table}%

\begin{table}[htbp]
  \centering
  \caption{Ablation studies in the object detection task. ``w/o'' denotes ``without'',  and ``GS'' indicates the globally accumulated statistic, \ie, $\hat{\mu}$ and $\hat{\sigma}$ in Eq.~\ref{eq07}.}
  \setlength{\tabcolsep}{3.3mm}{
    \begin{tabular}{l|cccc|c}
    \toprule
    \multicolumn{1}{c|}{Method} & NS & DR & NR & DF & Avg \\
    \midrule
    Faster R-CNN (FR) & 46.8  & 37.5  & 16.7  & 38.1  & 34.8 \\
    FR+PBN w/o GS & \textbf{47.0}  & 43.7  & 23.0  & 40.1  & 38.5 \\
    \rowcolor{gray!20} % 使用gray!20指定灰色的底色，20表示灰度值
    FR+PBN & \textbf{47.0}  & \textbf{44.3}  & \textbf{23.8}  & \textbf{40.4}  & \textbf{38.9} \\
    \bottomrule
     YOLOV5 & 33.5 & 25.5 & 10.4 & 38.1 & 26.9\\
     \rowcolor{gray!20}
    YOLOV5+PBN & \textbf{35.9} & \textbf{27.2} & \textbf{13.9} & \textbf{43.0} & \textbf{30.0} \\
      \bottomrule
    \end{tabular}}
  \label{tab07}%
   \vspace{-10pt}
\end{table}%

\begin{table*}%[t]
  \centering
  \caption{Comparison with some SOTA methods in the instance retrieval task. We run experiment three times and report the averaged result and standard deviation on Market1501 and GRID, respectively.}
    \setlength{\tabcolsep}{4.0mm}{
    \begin{tabular}{l|cccc|cccc}
    \toprule %\multicolumn{1}{c|}{Method}  \multirow{2}[1]{*}{Method}
       \multicolumn{1}{c|}{\multirow{2}[1]{*}{Method}} & \multicolumn{4}{c|}{Market1501$\to$GRID} & \multicolumn{4}{c}{GRID$\to$Market1501} \\
\cmidrule{2-9}    \multicolumn{1}{c|}{} & mAP   & R1    & R5    & R10   & mAP   & R1    & R5    & R10 \\
    \midrule
    Baseline & 33.3$\pm$0.4 & 24.5$\pm$0.4 & 42.1$\pm$1.0 & 48.8$\pm$0.7 & 3.9$\pm$0.4 & 13.1$\pm$1.0 & 25.3$\pm$2.2 & 31.7$\pm$2.0 \\
    %OSNet+PBN & 35.7$\pm$2.3 & 26.9$\pm$2.6 & 45.9$\pm$2.9 & 53.6$\pm$1.9 & 4.4$\pm$0.4 & 13.5$\pm$1.0 & 25.4$\pm$2.2 & 32.5$\pm$2.1 \\
    \midrule
    MixStyle~\cite{DBLP:conf/iclr/ZhouY0X21} & 33.8$\pm$0.9 & 24.8$\pm$1.6 & \textbf{43.7$\pm$2.0} & 53.1$\pm$1.6 & 4.9$\pm$0.2 & 15.4$\pm$1.2 & 28.4$\pm$1.3 & 35.7$\pm$0.9 \\
    \rowcolor{gray!20} % 使用gray!20指定灰色的底色，20表示灰度值
    MixStyle+PBN & \textbf{35.9$\pm$2.3} & \textbf{28.0$\pm$2.3} & 43.2$\pm$4.2 & \textbf{52.5$\pm$1.3} & \textbf{5.6$\pm$0.7} & \textbf{16.9$\pm$1.6} & \textbf{30.5$\pm$3.1} & \textbf{37.8$\pm$3.4} \\
    \midrule
    EFDMix~\cite{DBLP:conf/cvpr/ZhangLLJZ22} & 35.5$\pm$1.8 & 26.7$\pm$3.3 & 44.4$\pm$0.8 & \textbf{53.6$\pm$2.0} & 6.4$\pm$0.2 & 19.9$\pm$0.6 & 34.4$\pm$1.0 & 42.2$\pm$0.8 \\
    \rowcolor{gray!20} % 使用gray!20指定灰色的底色，20表示灰度值
    EFDMix + PBN & \textbf{37.0$\pm$0.9} & \textbf{28.0$\pm$2.0} & \textbf{44.8$\pm$0.7} & \textbf{53.6$\pm$1.1} & \textbf{6.8$\pm$0.3} & \textbf{20.5$\pm$0.5} & \textbf{35.4$\pm$0.9} & \textbf{43.1$\pm$0.8} \\
    \bottomrule
    \end{tabular}}
  \label{tab08}%
  %\vspace{-10pt}
\end{table*}%
% On Diverse-Weather, we train the model using the daytime-sunny domain, and test the trained model on other domains. Besides, unlike the classification task, there are some small objects in the object detection task, we keep the BN in the last block of ResNet and replace other BNs with PBN. These SOTA methods are implemented using avalibale codes provided by their authors, meanwhile our method is implemented based these codes.
\textbf{Results:}
In this section, we compare our patch-aware batch normalization with some recently proposed state-of-the-art single-domain generalized object detection methods, such as CycConf~\cite{DBLP:conf/iccv/0066HLY0GD21} and CDSD~\cite{DBLP:conf/cvpr/WuD22}. The experimental results are shown in Tab.~\ref{tab06}. As can be seen, using our method consistently improves the performance of SOTA methods. It is also worth noting that Faster R-CNN~\cite{DBLP:conf/nips/RenHGS15} is implemented using the Detectron2 \cite{wu2019detectron2} library, which includes the FPN module~\cite{DBLP:conf/cvpr/LinDGHHB17}. Thus, Faster R-CNN achieves better results than other methods.

% In this part, we compare our method with some recently SOTA single-domain generalized objection detection methods, such as CycConf~\cite{DBLP:conf/iccv/0066HLY0GD21} and CDSD~\cite{DBLP:conf/cvpr/WuD22}. The experimental results are listed in Table~\ref{tab06}. As seen in this table, using our method can consistently improve SOTA methods. In addition, it is worth noting that Faster R-CNN~\cite{DBLP:conf/nips/RenHGS15} is implemented using Detectron2 \cite{wu2019detectron2} library, which includes the FPN module~\cite{DBLP:conf/cvpr/LinDGHHB17}, thus Faster R-CNN can obtain better result than other methods. 

%\subsubsection{Ablation Studies}
We also conduct experiments on the object detection task to validate the effectiveness of the module in our PBN, as reported in Tab.~\ref{tab07}. Conducting patch-aware batch normalization (\ie, ``FR+PBN w/o GS'' in Tab.~\ref{tab07}) can improve the performance by $+3.7\%$ (38.5 vs. 34.8) compared to Faster R-CNN. Moreover, our method can be further improved by combining the globally accumulated statistics with the patch-wise statistics (\ie, ``FR+PBN'' in Tab.~\ref{tab07}). Besides, we integrate our PBN into YOLOv5s, as shown in Tab.~\ref{tab07}, indicate that using our PBN in YOLO can enhance the original generalization performance of YOLO.

% We also conduct the experiment in the object detention task to validate the effectiveness of the module in our PBN, as reported in Table~\ref{tab07}. Using the statistics of each patch can improve $+3.7\%$ (38.5 vs. 34.8) against Faster R-CNN. Moreover, combining with the globally accumulated statistics, our method can further be improved. 

\subsection{Generalization on Instance Retrieval}
%\textbf{Dataset:}
% We conduct the experiment in an instance retrieval task, called Person Re-identification (Re-ID), which aims at learning the generalizable features to match these images of the same identity from different cameras. We adopt the person
% Re-ID datasets, such as Markert1501~\cite{DBLP:conf/iccv/ZhengSTWWT15} and
% GRID~\cite{DBLP:conf/cvpr/LoyXG09}, to conduct cross-domain instance retrieval. We
% follow~\cite{DBLP:conf/cvpr/ZhangLLJZ22} to conduct experiments with the OSNet~\cite{DBLP:conf/iccv/ZhouYCX19}.
% Similar to the findings in  classification and object detection, when adding our PBN into existing SOTA methods, their performance can be consistently improved, as shown in Table~\ref{tab08}.
% This once again validates the effectiveness of utilizing the statistics based on patches to mitigate the overfitting to the training set in the cross-domain task.
We conduct an experiment on an instance retrieval task, namely Person Re-identification (Re-ID), which aims to learn generalizable features for matching images of the same identity from different cameras~\cite{Qi_2019_ICCV}. We use Person Re-ID datasets, such as Market1501~\cite{DBLP:conf/iccv/ZhengSTWWT15} and GRID~\cite{DBLP:conf/cvpr/LoyXG09}, to conduct cross-domain instance retrieval. We employ OSNet~\cite{DBLP:conf/iccv/ZhouYCX19} as the backbone and follow the experiment protocol of~\cite{DBLP:conf/cvpr/ZhangLLJZ22}. Similar to our findings in classification and object detection tasks, when our PBN was added to existing SOTA methods, their performance consistently improved, as shown in Tab.~\ref{tab08}. This further validates the effectiveness of using our PBN to mitigate the overfitting to the training set in cross-domain tasks.


\begin{table}[h]
  \centering
  \caption{Experimental results of the model trained using Cityscapes on BDD-100K (B), Mapillary (M), GTAV (G) and SYNTHIA (S) in the semantic segmentation task.}
    \setlength{\tabcolsep}{3.3mm}{
    \begin{tabular}{l|cccc|c}
    \toprule
    \multicolumn{1}{c|}{Method} & B & M & G & S & Avg \\
    \midrule
    Baseline & 44.5  & 53.0  & 41.4  & 24.3  & 40.8 \\
     \midrule
    RobustNet~\cite{DBLP:conf/cvpr/ChoiJYKKC21}   & 48.7  & 57.0  & 44.4  & 25.7  & 44.0 \\
    \rowcolor{gray!20} % 使用gray!20指定灰色的底色，20表示灰度值
    RobustNet+PBN & \textbf{50.6}  & \textbf{57.8}  & \textbf{46.6}  & \textbf{26.2}  & \textbf{45.3} \\
    \bottomrule
    \end{tabular}}
  \label{tab09}%
 % \vspace{-10pt}
\end{table}%

\subsection{Generalization on Semantic Segmentation}
%\textbf{Dataset:} 
In the semantic segmentation task, we conduct the multi-domain generalization experiment as in~\cite{DBLP:conf/cvpr/ChoiJYKKC21}. We train all methods on the Cityscapes dataset~\cite{cordts2016cityscapes}, which consists of multiple domains, and show their performance on other datasets, including BDD-100K~\cite{yu2020bdd100k}, Mapillary~\cite{neuhold2017mapillary}, GTAV~\cite{richter2016playing}, and SYNTHIA~\cite{ros2016synthia}, to measure the generalization capability on unseen domains.

In this section, we adopt DeepLabV3+~\cite{DBLP:conf/eccv/ChenZPSA18} as the backbone, following the setting in~\cite{DBLP:conf/cvpr/ChoiJYKKC21}. We compare our method with RobustNet~\cite{DBLP:conf/cvpr/ChoiJYKKC21}, which disentangles the domain-specific style and domain-invariant content encoded in higher-order statistics (\ie, feature covariance) of the feature representations and selectively removes only the style information causing domain shift. The experimental results are reported in Tab.~\ref{tab09}. As can be seen in the table, RobustNet improves the performance of the Baseline, and plugging our PBN into the RobustNet can further generate better results. Additionally, we conduct an ablation study in the semantic segmentation task, as shown in Tab.~\ref{tab10}, which validates the efficacy of using the patch-aware normalization and globally accumulated statistics.

\begin{table}[h]%[htbp]
  \centering
  \caption{Ablation studies in semantic segmentation. ``C'' denotes the training set, \ie, Cityscapes. ``w/o'' denotes ``without'',  and ``GS'' is the globally accumulated statistic, \ie, $\hat{\mu}$ and $\hat{\sigma}$ in Eq.~\ref{eq07}.}
  \setlength{\tabcolsep}{3.2mm}{
    \begin{tabular}{l|cccc|c}
    \toprule
     \multicolumn{1}{c|}{Method} & B & M & G & S & Avg \\
    \midrule
    Baseline (BL) & 44.5  & 53.0  & 41.4  & 24.3  & 40.8 \\
    BL+PBN w/o GS & \textbf{46.2}  & 53.7  & \textbf{43.5}  & 25.3  & 42.2 \\
    \rowcolor{gray!20} % 使用gray!20指定灰色的底色，20表示灰度值
    BL+PBN & 46.1  & \textbf{57.2}  & 42.7  & \textbf{25.4}  & \textbf{42.9} \\
    \bottomrule
    \end{tabular}}
  \label{tab10}%
 % \vspace{-10pt}
\end{table}%

% \begin{figure}[t]
%   \centering
%   \begin{subfigure}{1.0\linewidth}
%   \includegraphics[width=8cm]{fig/fig_s1.pdf}%2.95
%       \caption{\scriptsize{Images for classification.}}
%     \label{fig:short-a}
%   \end{subfigure}
%   \hfill
%   \begin{subfigure}{1.0\linewidth}
%      \includegraphics[width=8.5cm]{fig/fig_s2.pdf}
%     \caption{\scriptsize{Images for detection and segmentation.}}
%     \label{fig:short-b}
%   \end{subfigure}
% \caption{Visualization of differences between non-overlapping patches within an image. In this figure, ``lu'', ``ru'', ``ld'' and ``rd'' indicate the position of each patch, where ``lu'' stands for the left-upper patch, ``ru'' for the right-upper patch, "ld'' for the left-down patch, and ``rd'' for the right-down patch. Each curve in the figure represents mean or standard deviation of a patch. ``global'' indicates the global statistics. The second and third rows represent the mean ($\mu$) and standard deviation ($\sigma$), respectively.}\label{figs1}
% \vspace{-20pt}
% \end{figure}

\begin{figure}[!h]
\centering
\subfigure[Images for classification.]{
\begin{minipage}[b]{0.95\linewidth}
\centering
\includegraphics[width=\textwidth]{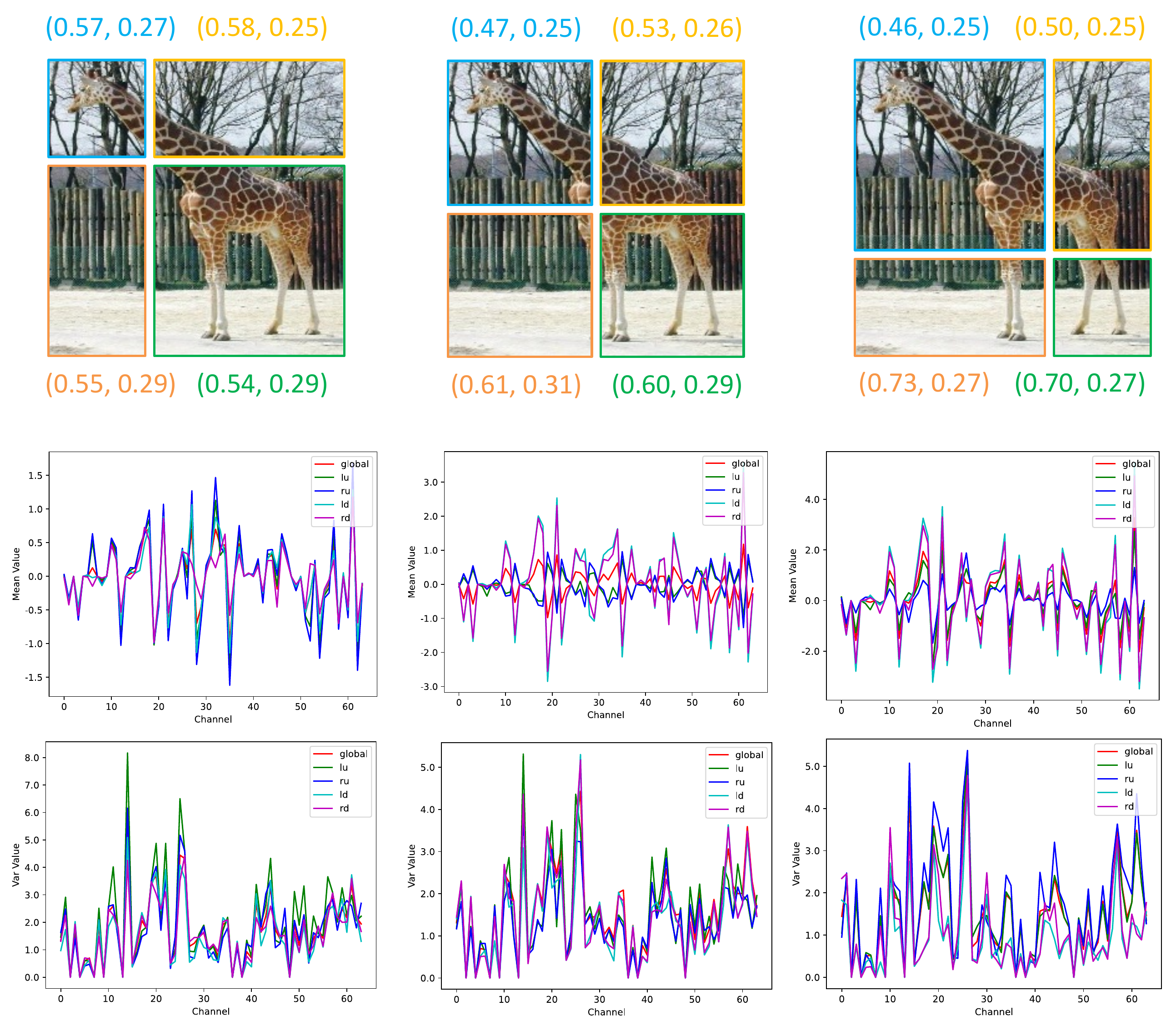}
\end{minipage}
}
\subfigure[Images for detection and segmentation.]{
\begin{minipage}[b]{0.95\linewidth}
\centering
\includegraphics[width=\textwidth]{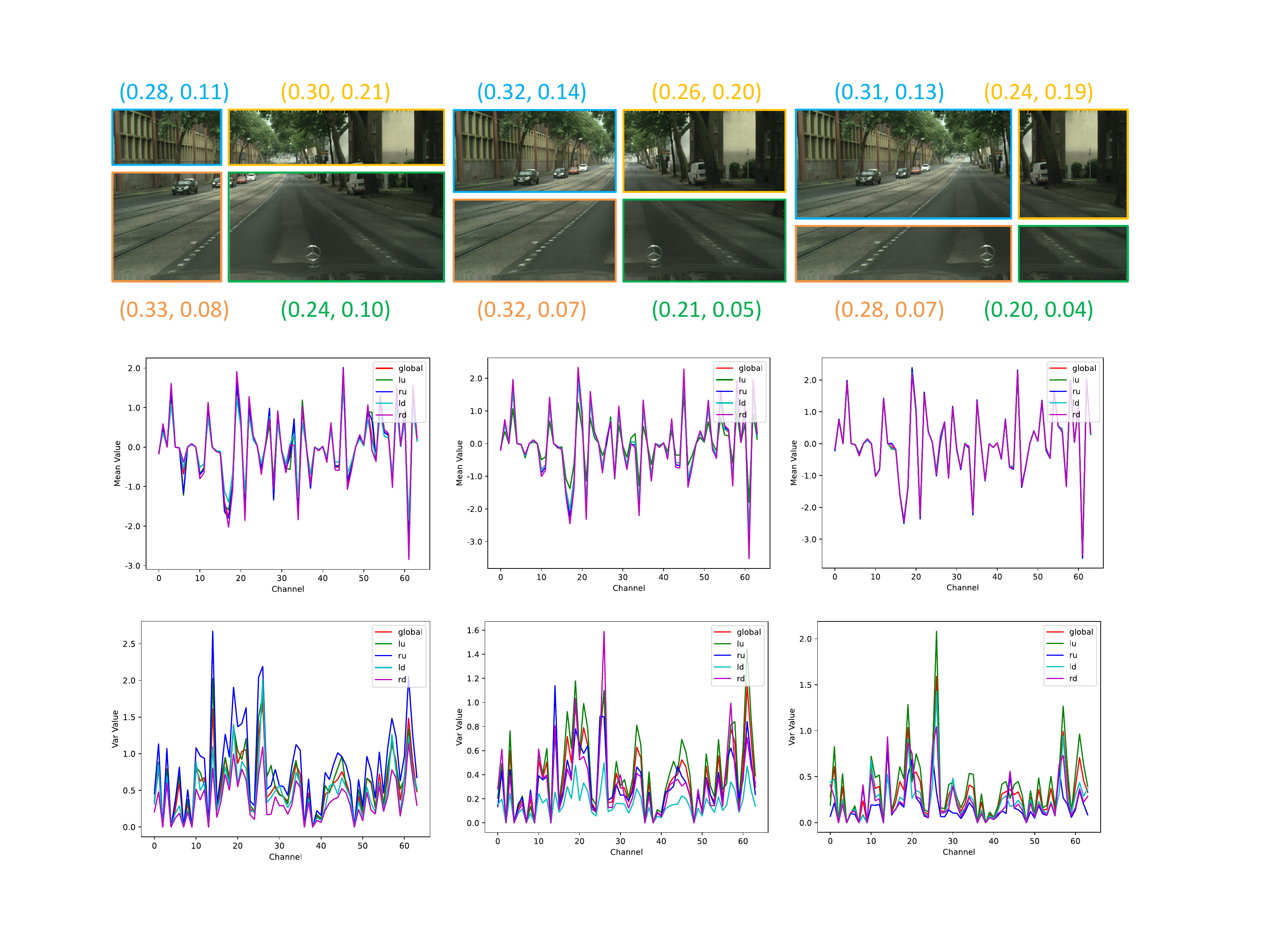}
\end{minipage}
}
\caption{Visualization of differences between non-overlapping patches within an image. In this figure, ``lu'', ``ru'', ``ld'' and ``rd'' indicate the position of each patch, where ``lu'' stands for the left-upper patch, ``ru'' for the right-upper patch, "ld'' for the left-down patch, and ``rd'' for the right-down patch. Each curve in the figure represents mean or standard deviation of a patch. ``global'' indicates the global statistics. The second and third rows are the mean ($\mu$) and standard deviation ($\sigma$).}
\label{figs1}
\vspace*{-10pt}
\end{figure}

\subsection{Further Analysis}

\subsubsection{Visualization of Statistics}
We have generated a visualization of the differences between local patches within an image, as shown in Fig.~\ref{figs1}. To obtain this visualization, we compute the statistics of feature maps before the first batch normalization (BN) layer of ResNet~\cite{DBLP:conf/cvpr/HeZRS16}, resulting in $(\mu, \sigma)$ where $\mu$ and $\sigma$ are vectors of length $C$ (\ie, the number of channels), and in this case, $C$ is 64. As can be seen in Fig.~\ref{figs1} (b), there are obvious differences in the standard deviations between the different patches, while the mean values are only slightly different. We also show that different partitions with varying sizes can introduce additional discrepancies, which is beneficial for mitigating the impact of overfitting.

% \begin{wrapfigure}{r}{0.29\linewidth}
%   \vspace{-0.3cm}
%   % \certering
%   \hspace{-0.6cm}
%   \includegraphics[scale=0.32]{fig/figR3.pdf}
%   \vspace{-0.3cm}
% \end{wrapfigure}

% \begin{figure}[t]
% \begin{center}
%    \includegraphics[width=0.9\linewidth]{fig/figR3.pdf}
%    \caption{The experimental results of our PBN on VGG on the PACS dataset.}
% \label{fig03}
% \end{center}
% \vspace{-25pt}
% \end{figure}
  \setcounter{table}{12}
\begin{table}%[h]%[htbp]
  \centering
  \caption{The experimental results of our PBN on VGG on the PACS dataset.}
  \setlength{\tabcolsep}{3.2mm}{
    \begin{tabular}{l|C{4.5mm}C{4.5mm}C{4.5mm}C{4.5mm}|c}
    \toprule
     \multicolumn{1}{c|}{Method} & P     & A     & C     & S     & Avg \\
    \midrule
    Baseline(BL) & 60.4 & 63.5 & 38.9 & 28.6 & 47.8 \\
    BL+PBN w/o GS & \textbf{60.9} & 65.6 & 39.8 & 32.1 & 49.6 \\
    \rowcolor{gray!20} % 使用gray!20指定灰色的底色，20表示灰度值
    BL+PBN & 60.0 & \textbf{66.0} & \textbf{39.9} & \textbf{33.6} & \textbf{49.9} \\
    \bottomrule
    \end{tabular}}
  \label{tab:vgg}%
 % \vspace{-5pt}
\end{table}%

% \begin{table}[h]
%    \vspace{-10pt}
%    \begin{center}
%       \renewcommand\arraystretch{1.0}
%       % \caption{Mixture of accumulated statistics $\alpha * \text{tgt} + (1-\alpha)*\text{src}$.}
%       \resizebox{1\columnwidth}{!}{
%     \begin{tabular}{cc||cc||ccc}
%     \toprule
%     ADA & {\cellcolor[rgb]{ .851,  .851,  .851}ADA+PBN} & ME-ADA & {\cellcolor[rgb]{ .851,  .851,  .851}ME-ADA+PBN} & BL & {\cellcolor[rgb]{ .988,  .894,  .839}BL+PBN w/o GS} & {\cellcolor[rgb]{ .988,  .894,  .839}BL+PBN} \\
%     \midrule
%     69.8  & \cellcolor[rgb]{ .851,  .851,  .851}\textbf{77.0} & 76.9  & \cellcolor[rgb]{ .851,  .851,  .851}\textbf{82.4} & 69.3  & \cellcolor[rgb]{ .988,  .894,  .839}74.3 & \cellcolor[rgb]{ .988,  .894,  .839}\textbf{74.9} \\
%     \bottomrule
%     \end{tabular}%
%          \label{mix}
%       }
%    \end{center}
%    \vspace{-15pt}
% \end{table}

% Table generated by Excel2LaTeX from sheet 'rebuttal'
%\begin{table}[htbp]

\begin{table}[!h]%[htbp]
  \centering
  \caption{Performance assessment on the PACS dataset: leveraging a model founded on the \underline{{\color{newblue}ViT}} architecture.}
  \setlength{\tabcolsep}{3.5mm}{
    \begin{tabular}{l|cccc|c}
    \toprule
     \multicolumn{1}{c|}{Method} & P     & A     & C     & S     & Avg \\
    \midrule
    ViT-small & 64.5 & 64.7 & 39.8 & 50.9 & 55.0 \\
    \rowcolor{gray!20} % 使用gray!20指定灰色的底色，20表示灰度值
    ViT-small+PLN & \textbf{66.3} & \textbf{65.9} & \textbf{44.2} & \textbf{50.9} & \textbf{56.8} \\
    \midrule
    ViT-large & 72.5 & 74.8 & 48.1 & 75.8 & 67.8 \\
    \rowcolor{gray!20} % 使用gray!20指定灰色的底色，20表示灰度值
    ViT-large+PLN & \textbf{73.4} & \textbf{77.2} & \textbf{48.2} & \textbf{78.3} & \textbf{69.3} \\
    \bottomrule
    \end{tabular}}
  \label{03}%
 % \vspace{-10pt}
\end{table}%

% \setlength{\columnsep}{10pt}
% \begin{wraptable}{r}{2.7cm}
% \vspace{-1em}
% \tiny
% \centering
%  \setlength{\tabcolsep}{0.5mm}{
% \begin{tabular}{l|c|c} \toprule
% \multicolumn{1}{c|}{Method} & \multicolumn{1}{c|}{Source} & \multicolumn{1}{c}{Cross} \\ \arrayrulecolor[rgb]{0.502,0.502,0.502}\hline
% ViT-small & \textbf{99.5} & 55.0 \\
% \rowcolor[rgb]{0.847,0.847,0.847} ViT-small w PLN & \textbf{99.5} & \textbf{56.8} \\ \hline
% ViT-large & 99.7 & 67.8 \\
% \rowcolor[rgb]{0.847,0.847,0.847} ViT-large w PLN & \textbf{99.8} & \textbf{69.3} \\ \bottomrule
% \end{tabular}
% }
% \vspace{-2em}
% \caption{Performance assessment on the PACS dataset: leveraging a model founded on the \underline{{\color{newblue}ViT}} architecture.}
% \vspace{-3em}
% \label{03}
% \end{wraptable}

\subsubsection{Comparison to those BN, LN, IN and GN}
In Tab.~\ref{norm}, we compare our method with other typical normalizations. As observed, our PBN achieves superior performance compared to the alternatives. It is worth noting that IN shows poor performance due to its limitation in eliminating discriminative patterns that might be beneficial for a specific instance, as stated in the abstract of the literature.

\newcommand{\myTable}
{
  \centering
   \caption{Experimental results on CIFAR-10 based on DenseNet.}
     \resizebox{0.2\textwidth}{!}{
    \begin{tabular}{l|c}
    \toprule
    \multicolumn{1}{c|}{Method} & Accuracy \\
    \midrule
    ADA~\cite{DBLP:conf/nips/VolpiNSDMS18}    & 69.8 \\
    \rowcolor{gray!20} % 使用gray!20指定灰色的底色，20表示灰度值
    ADA+PBN & \textbf{77.0} \\
    \midrule
    ME-ADA~\cite{DBLP:conf/nips/000300M20} & 76.9 \\
    \rowcolor{gray!20} % 使用gray!20指定灰色的底色，20表示灰度值
     ME-ADA\newline{}+PBN & \textbf{82.4} \\
    \midrule
    BL    & 69.3 \\
    \midrule
    BL+PBN \newline{}w/o GS & 74.3 \\
    \rowcolor{gray!20} % 使用gray!20指定灰色的底色，20表示灰度值
    BL+PBN & \textbf{74.9} \\
    \bottomrule
    \end{tabular}%
    }
    }
  \setcounter{table}{11}
\begin{wraptable}{l}{4.5cm}
\myTable
\label{tabS01}%
\end{wraptable}
\subsubsection{Experiments based on ViT}
We extend our method to Layer Normalization (referred to as PLN) using the ViT and present the experimental results in Tab.~\ref{03}. Additionally, we have partitioned the patch dimension of the features using the P4-random method. As evident, this extension also contributes to an improvement in the performance of ViT.

\begin{table}[]%[htbp]
  \centering
  \caption{Experiments on the PACS Dataset with previous normalization methods.}
  % \vspace{-10pt}
  \setlength{\tabcolsep}{4.2mm}{
    \begin{tabular}{l|cccc|c}
    \toprule
     \multicolumn{1}{c|}{Method} & P     & A     & C     & S     & Avg \\
    \midrule
    BN & 58.6 & 66.4 & 34.0 & 27.5 & 46.6 \\
    IN & 23.8 & 16.3 & 17.2 & 11.8 & 17.3 \\
    LN & 54.4 & 48.4 & 20.5 & 28.4 & 37.9 \\
    GN & 58.0 & 52.6 & 32.9 & \textbf{44.9} & 47.1 \\
    \midrule
    \rowcolor{gray!20} % 使用gray!20指定灰色的底色，20表示灰度值
    PBN & \textbf{60.5} & \textbf{69.1} & \textbf{36.2} & 31.5 & \textbf{49.3} \\
    \bottomrule
    \end{tabular}}
  \label{norm}%
  \vspace{-20pt}
\end{table}%

\subsubsection{Experiments based on Other Backbones}
We also add the experiment based on \underline{VGG}~\cite{simonyan2015very} on PACS in Tab.~\ref{tab:vgg} and \underline{DenseNet}~\cite{huang2017densely} on CIFAR-10 in Tab.~\ref{tabS01}, respectively. As observed in the tables, our PBN consistently demonstrates its effectiveness with different backbones, which further confirm the efficacy of our Patch-aware Batch Normalization.

  \setcounter{table}{15}
\begin{table}[htbp]
  \centering
  \caption{Combining PBN with various augmentation methods on the PACS dataset to validate the stability and scalability of PBN}
  \setlength{\tabcolsep}{3.0mm}{
    \begin{tabular}{l|cccc|c}
    \toprule
     \multicolumn{1}{c|}{Method} & P     & A     & C     & S     & Avg \\
    \midrule
    Strong\_Aug & 61.3 & 70.8 & 37.6 & 40.9 & 52.7 \\
    \rowcolor{gray!20} % 使用gray!20指定灰色的底色，20表示灰度值
    PBN w Strong\_Aug & \textbf{63.5} & \textbf{73.0} & \textbf{38.5} & \textbf{42.2} & \textbf{54.3} \\
    \midrule
    RandAug & 65.3 & 71.1 & 42.0 & 49.0 & 56.8 \\
    \rowcolor{gray!20} % 使用gray!20指定灰色的底色，20表示灰度值
    PBN w RandAug & \textbf{68.9} & \textbf{72.5} & \textbf{44.1} & \textbf{50.0} & \textbf{58.9} \\
    \midrule
    AutoAug & 61.0 & 71.2 & 40.3 & 54.9 & 56.9 \\
    \rowcolor{gray!20} % 使用gray!20指定灰色的底色，20表示灰度值
    PBN w AutoAug & \textbf{64.8} & \textbf{72.1} & \textbf{42.0} & \textbf{55.7} & \textbf{58.6} \\
    \bottomrule
    \end{tabular}}
  \label{aug}%
  \vspace{-5pt}
\end{table}%

\subsubsection{The stability and sensitivity of PBN}
Based on different augmentations, we use PBN for experiment, as shown in Tab.~\ref{aug}. Our PBN still keeps the stable improvement.

\subsubsection{Experiments on the larger-scale datasets}
%We extend our validation of the effectiveness of the PBN to larger-scale datasets, including ImageNet-C, ImageNet-A, and Stylized-ImageNet, as presented in Tab.~\ref{04}. As evident, our PBN outperforms the model without PBN. The aforementioned experimental results convincingly establish the applicability of PBN on large-scale datasets.
We extend our validation of the effectiveness of the PBN to larger-scale datasets, including ImageNet-C, ImageNet-A, and Stylized-ImageNet (Sty-IN), as presented in Tab.~\ref{04}. Across these diverse and challenging datasets, our PBN consistently outperforms the baseline model without PBN, demonstrating its robustness and generalization capabilities. The improvements in performance are particularly noteworthy on ImageNet-C and ImageNet-A, which contain various corruptions and natural adversarial examples, respectively. This highlights the ability of PBN to enhance model robustness against distribution shifts and adversarial perturbations. Additionally, the promising results on Stylized-ImageNet, which features stylized renditions of natural images, showcase the effectiveness of PBN in handling diverse visual domains. The aforementioned experimental results convincingly establish the applicability and versatility of PBN on large-scale datasets, solidifying its potential for widespread adoption in real-world computer vision applications.
\begin{table}%[h]
\centering
\caption{Experimental evaluation on ImageNet (ImN). ImageNet is source domain, and we test the model in different domains. It is worth noting that for mCE, the smaller value is better; for Top-1 ACC, the larger value is better. ``R50'' is the ResNet-50.}
\vspace{-10pt}
%\arrayrulecolor[rgb]{0.502,0.502,0.502}
\begin{center}
      %\renewcommand\arraystretch{1.0}
      % \caption{Mixture of accumulated statistics $\alpha * \text{tgt} + (1-\alpha)*\text{src}$.}
      % \resizebox{1\columnwidth}{!}{
\setlength{\tabcolsep}{3.2mm}{
\begin{tabular}{l|c|c} \toprule
\multicolumn{1}{c|}{Method} & \multicolumn{1}{c|}{ImN-C (mCE)} & \multicolumn{1}{c}{ImN-A (Top-1 Acc)}  \\ \hline
R50 & 74.8 & 3.1  \\
\rowcolor[rgb]{0.847,0.847,0.847} R50+PBN & \textbf{68.3 $\downarrow$\textcolor{morandired}{6.5}} & \textbf{6.1 $\uparrow$\textcolor{morandired}{3.0}} 
 \\ \hline
R50+AuxBN\cite{DBLP:conf/eccv/JiaoLGLYZWZ22} & 69.7 & 4.8  \\
\rowcolor[rgb]{0.847,0.847,0.847} R50+AuxBN+PBN & \textbf{66.5 $\downarrow$\textcolor{morandired}{3.2}} & \textbf{7.0 $\uparrow$\textcolor{morandired}{2.2}} 
 \\ \bottomrule
 \multicolumn{1}{c|}{Method} & \multicolumn{1}{c|}{Sty-ImN (Top-1 Acc)} & \multicolumn{1}{c}{ImN (Top-1 Acc)}  \\\hline
R50  & 8.0 & 76.6 \\
\rowcolor[rgb]{0.847,0.847,0.847} R50+PBN & \textbf{11.6 $\uparrow$\textcolor{morandired}{3.6}} & \textbf{77.7 $\uparrow$\textcolor{morandired}{1.1}}
 \\ \hline
R50+AuxBN\cite{DBLP:conf/eccv/JiaoLGLYZWZ22}  & 10.3 & 77.5 \\
\rowcolor[rgb]{0.847,0.847,0.847} R50+AuxBN+PBN & \textbf{13.4 $\uparrow$\textcolor{morandired}{3.1}} & \textbf{77.9 $\uparrow$\textcolor{morandired}{0.4}}
 \\ \bottomrule
\end{tabular}}
\label{04}
\end{center}
\vspace{-5pt}
\end{table}

\subsubsection{Evaluation on the training time.}
We measure the time required to train for one epoch using the original baseline method and our integrated method, as shown in Tab.~\ref{tabR5}. Our proposed method only slightly increases the training time compared to the baseline. This is because our method involves new statistical calculations only in the BN layer and does not affect other layers in the backbone.
\begin{table}[h]
  \centering
  \caption{The training times (second) of our method.}
      \setlength{\tabcolsep}{10.0mm}{
      \vspace{-10pt}
    \begin{tabular}{l|c}
    \toprule
    \multicolumn{1}{c|}{Method}  & Time/Epoch \\
    \midrule
    ResNet18   & 5.0483 \\
   \rowcolor[rgb]{0.847,0.847,0.847} PBN-ResNet18 & 5.1738 \\
    \midrule
    ResNet50   & 9.3048 \\
   \rowcolor[rgb]{0.847,0.847,0.847} PBN-ResNet50 & 9.5886 \\
    \bottomrule
    \end{tabular}%
    }
  \label{tabR5}
\end{table}%

\subsubsection{Comparison with Meta-BIN.}
We compare our proposed method with Meta-BIN in the DG-ReID setting, and our method can achieve better performance, as shown in Tab.~\ref{tabR8}. Compared to Meta-BIN, our method can bring more diverse information.

\begin{table}[h]
  \centering
  \caption{Comparison between MetaBIN and our PBN in the DG-ReID setting.}
  % \vspace{-10pt}
    \begin{tabular}{l|cc|cc}
    \toprule
  \multicolumn{1}{c|}{\multirow{2}[1]{*}{Method}}  & \multicolumn{2}{c|}{M$\to$G} & \multicolumn{2}{c}{G$\to$M} \\
\cmidrule{2-5}     & mAP & R1 & mAP & R1 \\
    \midrule
    MetaBIN~\cite{choi2021meta} & 34.05 & 24.84 & 4.26  & 13.45 \\
    PBN   & \textbf{35.67} & \textbf{26.93} & \textbf{4.43} & \textbf{13.53} \\
    \bottomrule
    \end{tabular}%
  \label{tabR8}%
\end{table}%

\section{Conclusion}\label{s-conclusion}
%In conclusion, this paper addresses the challenges of single-source domain generalization by the proposed CPerb, a cross-perturbation method. By leveraging both horizontal and vertical operations, CPerb effectively enhances the ability of the model to generalize to unknown domains. The horizontal operations, which include image- and feature-level perturbations, alleviate the issue of limited diversity in the training data, while the vertical operations, employing multi-route perturbation, facilitate the learning of domain-invariant features. Additionally, the introduction of MixPatch, a novel feature-level perturbation method, further enhances the diversity of the training data by exploiting local image style information. Experimental results on various benchmark validate the effectiveness of our method. The proposed CPerb framework shows promise in improving the generalization performance of models in single domain generalization tasks. Future work can explore additional techniques to further enhance the capabilities of CPerb and apply it to other related domains.
In this paper, we propose a novel method called patch-aware batch normalization (PBN) for enhancing cross-domain robustness by exploiting the differences between local patches within an image to alleviate overfitting issues. PBN divides feature maps into different patches and performs normalization operations independently, leading to better optimization of the BN parameters. Additionally, globally accumulated statistics are integrated to mitigate the inaccuracy of patch statistics. Multiple experiments on different tasks validate the effectiveness of our proposed PBN.

In our experiments, we also attempt to apply our method to the multi-domain generalization classification task, but it does not result in significant improvements. In contrast to the cityscapes dataset, there may be significant differences across multiple domains in the multi-domain generalized classification task, which can introduce diverse information when randomly selecting images to form each batch. For instance, on the PACS dataset, where we use Photo (P), Art (A), and Sketch (S) to train the model, the domain gap is large in the training set. Using our method in this scenario could introduce more randomness that may act as noise during training, because we observe that using our PBN leads to a larger standard deviation than the baseline. We will investigate this issue in future work.



%\section*{Acknowledgment}

\ifCLASSOPTIONcaptionsoff
  \newpage
\fi

\bibliographystyle{IEEEtran}
\bibliography{sigproc}

\end{document}